\documentclass{article}
\usepackage[preprint]{tmlr}

\usepackage{amsmath,amsfonts,bm}

\def\eqref#1{equation~\ref{#1}}
\def\1{\bm{1}}

\DeclareMathAlphabet{\mathsfit}{\encodingdefault}{\sfdefault}{m}{sl}
\SetMathAlphabet{\mathsfit}{bold}{\encodingdefault}{\sfdefault}{bx}{n}

\usepackage{amsmath,amssymb}
\usepackage{booktabs}
\usepackage{multirow}
\usepackage{graphicx}
\usepackage{xcolor}
\usepackage{hyperref}
\hypersetup{colorlinks=true, allcolors=blue!50!black}
\usepackage{url}
\usepackage{placeins}
\usepackage{flafter}
\title{What Does Attention Transfer Transfer? \\ Attention Structure and Robustness in Vision Transformers}

\author{\name Jesse Ponnock \email jponnoc1@jh.edu \\
\addr Johns Hopkins University}

\def\month{08}
\def\year{2026}
\def\openreview{\url{https://openreview.net/forum?id=XXXX}}

\begin{document}

\maketitle

\begin{abstract}
Vision transformers (ViTs) trained to copy a pretrained teacher's attention maps recover most
of fine-tuning's
in-distribution accuracy yet fall measurably short of it under distribution shift, as recent
work has shown. What the copy delivers has never been measured directly in the attention
structure and tied to robustness. We build that instrumentation for ViT-S students of a
self-supervised teacher on ImageNet-100, and report three findings that triangulate one
conclusion. First, the transfer is essentially perfect and permanently so: the distilled
student's attention ends up roughly two orders of magnitude closer to the teacher's than
fine-tuning does, and does not drift with additional training. Second, the gap is real at
14$\times$ fewer parameters and 10$\times$ less data than previously studied, but it has a
time axis. It tracks
training maturity, and completing the schedules that the stopping rule interrupted closes it below our
pre-registered threshold in two of three seeds, with comparisons at equal accuracy giving the
same result. The endpoint gap at this scale is substantially a training-maturity artifact:
robustness matures later than accuracy, and stopping rules tuned to accuracy undersample it.
Third, forcing cross-row redundancy down by half the structural separation between the
distilled and fine-tuned conditions produces no
detectable robustness response under two registered ways of matching accuracy. Verified transfer, a gap that closes while the structure never
moves, and a null under direct intervention are together consistent with the deficit residing
in features, not in the visible attention structure. This is elimination plus intervention,
and its scope is the regime we measured. In this regime, attention overlays show where a model
looks, not what it knows.

\end{abstract}

\section{Introduction}\label{sec:intro}

Attention maps are among computer vision's most trusted windows into a transformer. They are
overlaid on images to show a model ``attending to the object''
\citep{wu2024faithfulness}, scored for quality
\citep{zhou2022understanding, darcet2024vision}, and distilled from teacher to student
\citep{zagoruyko2017paying} on the premise that they carry what the teacher knows. Yet a
student trained to copy a pretrained teacher's attention maps recovers most of fine-tuning's
clean accuracy while remaining measurably less robust to distribution shift
\citep{li2024attention}, a gap conjectured but never tested to lie in the features. If the
attention arrives intact, why doesn't the robustness?

This paper builds the instrumentation to answer that question, and the answer reframes it.
Attention transfer copies what an attention map shows, the routing that decides which tokens
read from which, while the student learns its own features from scratch. We
train scratch, fine-tuned, and attention-distilled ViT-S students (the latter two inheriting
from a single self-supervised teacher), and measure what attention transfer actually
transfers, for the first time directly in the attention structure and tied to robustness.
Three findings compose the answer.

First, \textbf{the copy is essentially perfect, and permanently so.} In per-layer, per-head
KL divergence on held-out images, the distilled student's attention sits on average 97$\times$ closer to
its teacher's than fine-tuning ends up from the same teacher, though fine-tuning began from
the teacher's own weights. Every
layer, in every seed, sits at least 30$\times$ closer than fine-tuning, and the
structure holds across stopping epochs spanning half the schedule and across a wide accuracy
range. Attention
transfer transfers attention. Whatever explains the gap, it is not a failed copy.

Second, \textbf{the gap is real but has a time axis.} It reproduces far below the scale of \citet{li2024attention},
at 14$\times$ fewer parameters and 10$\times$ less data, under the pre-registered rule
and horizons. Across all fifteen distill-family runs, endpoint effective robustness rises with
stopping epoch at $r = 0.978$. The gap shrinks as distillation approaches
schedule completion, while the attention structure never moves. What matures is not the
routing. Complete the schedules and the gap itself closes. Full-schedule reruns of all three
distilled seeds bring the gap below our pre-registered threshold of 0.01 in two of three, with
comparisons at equal accuracy giving the same result. The endpoint robustness gap at
this scale is substantially a training-maturity artifact.

Third, \textbf{structure responds to force; robustness does not follow.} Adding an attention
diversification penalty \citep{guo2023robustifying} at increasing dose moves the transferred
structure monotonically toward lower cross-row redundancy, by half the structural distance
between conditions.
Sharpness is untouched, corruption-stability moves an order of magnitude less, and
completed-schedule accuracy is uncosted. This is a causal dial on the one component everyone
can see. Turning it produces no detectable robustness response under either of two
registered ways of matching accuracy. The deficit is consistent with residing in
features, not in the visible attention structure.

\paragraph{Contributions.}
\begin{itemize}
\item The first direct measurement of what attention-map distillation does to attention
  structure, tied to robustness, covering transplant fidelity, maturity-invariance, and the
  two-senses dissociation of ``overfocusing,'' where concentration and redundancy move
  independently.
\item Reproduction of the gap under pre-registration, with its maturity structure and
  within-seed closure at completed schedules, revising how fixed-budget robustness comparisons
  should be read.
\item A causal dose-response on transferred attention structure, with a robustness null at
  matched skill under a displacement of half the condition separation.
\item A measurement methodology of matched-accuracy coordinates, stopping-variance
  quantification, and continuation calibration, reusable wherever early stopping meets
  no-peak curves.
\end{itemize}

\section{Related Work}\label{sec:related}

\paragraph{Attention transfer and what distillation carries.}
Transferring a teacher's attention is one of the oldest ideas in knowledge distillation
\citep{zagoruyko2017paying}. Its modern ViT form trains a randomly-initialized student to match
a pretrained teacher's attention maps. \citet{li2024attention} showed that such a student
recovers most of fine-tuning's in-distribution accuracy while falling measurably short on
distribution shift. They measured that gap at fixed budgets, with no stopping rule and no
convergence evidence, and conjectured but did not test it to be feature-borne. We build the instrumentation such a test
requires, and the budget condition turns out to matter: at our scale the endpoint gap is
substantially a training-maturity artifact (Section~\ref{sec:gap}). Li et al.'s own appendix makes a
first gesture in this direction (a layer-wise JSD between attention maps, with an acknowledged
identifiability caveat); we replace it with a purpose-built multi-axis panel tracked across
training, seeds, and doses, tied to robustness. A parallel line asks what distillation actually
transmits: \citet{stanton2021does} showed distillation can improve students whose predictions
nonetheless remain far from the teacher's; \citet{ojha2023knowledge} characterized what
knowledge survives the process; and concurrent work by \citet{qin2026attention} uses per-layer
student--teacher attention divergence as a diagnostic for when attention transfer fails across
architectures. We differ from all three in object and method: we measure the \emph{structure}
of what arrives (concentration, redundancy, stability) rather than agreement alone, show that
structure is transplanted essentially intact and held across training, and then intervene on it
causally. Where our results overlap with Qin et al.'s diagnosis, they agree: faithful attention
does not guarantee transferred competence. Their finding is convergent evidence for a
feature-borne account, from an independent design. Benchmarking of distillation under shift exists
\citep{zhang2025shiftkd} but does not study training-duration effects, which is where our result
lives.

\paragraph{Attention pathologies in ViTs.}
Two distinct notions of ``bad attention'' circulate, usually unseparated. One is per-row
collapse: attention entropy imploding as tokens fixate \citep{zhai2023stabilizing}. The other
is cross-row redundancy: every query attending to the same few tokens, identified by
\citet{guo2023robustifying}, whose Attention Diversification Loss we adopt as our intervention.
The ``overfocusing'' construct conflates these two measurably independent properties; no prior
work separates them. Our measurements show they dissociate across training
conditions and can be manipulated independently: ADL moves cross-row redundancy at dose without
touching per-row sharpness. That recasts the construct as two-dimensional, and suggests
earlier results attributed to ``overfocusing'' may need re-attribution to one sense or the
other. Related interventions encourage feature diversity for generalization
\citep{nicolicioiu2023diverse}; our use of ADL is diagnostic rather than performance-seeking.

\paragraph{Attention as evidence.}
NLP litigated whether attention weights explain predictions \citep{jain2019attention,
wiegreffe2019attention}; vision has since built faithfulness metrics for attention-derived
saliency \citep{wu2024faithfulness}. That entire thread concerns the
\emph{prediction-explanation} axis: does attention explain why this input produced this
output? Ours is a different axis: whether attention structure carries \emph{robustness}. We probe it
with a design missing from that thread, a transplant experiment in which the ``good-looking''
attention is held fixed by construction while the property it supposedly certifies fails to
arrive. The reading we counter is a live practice, not a strawman: attention quality is used as
evidence of model quality, most prominently where attention-map properties are credited for
robustness \citep{zhou2022understanding}, and artifact-free maps are engineered as desirable in
themselves \citep{darcet2024vision}. The overlays are not wrong about where the model looks;
they are uninformative, in our regime, about what survives distribution shift.

\paragraph{Robustness dynamics during training.}
Effective robustness \citep{taori2020measuring, miller2021accuracy} is known to \emph{evolve}.
Those works define a fitted-baseline residual; our ratio-form usage is defined and
distinguished in Section~\ref{sec:setup}.
\citet{andreassen2021evolution} showed that pretrained models' effective robustness erodes over
fine-tuning, and \citet{lin2024robustness} describe distributional overfitting, with OOD
performance deteriorating late in training while ID accuracy still creeps up. Adversarial
robustness shows its own overfitting dynamics \citep{rice2020overfitting}, and a rhyming
decay of teacher-alignment value with training maturity appears in diffusion training
\citep{wang2025haste}. In our distillation
regime the ratio-form quantity we report rises with training maturity ($r = 0.978$ against
stopping epoch), and the attention-transfer gap shrinks accordingly. We do not claim a sign
reversal against Andreassen et al., whose quantity is the fitted-baseline residual: under
that definition our maturity gradient itself reads negative
(Appendix~\ref{app:residual}); we report both directions. The lesson is not ``robustness always improves late'' but that
fixed-budget robustness comparisons are regime-dependent measurements of a moving quantity, and
can invert.

\paragraph{Run-to-run variance and underspecification.}
Our stopping-lottery finding extends a known theme: performance
variance from stopping and seeds \citep{dodge2020fine} and underspecification of training
pipelines \citep{damour2022underspecification}. One rule stopped same-condition runs 145 epochs
apart, differing only in seed, with endpoint metrics tracking the cut. Our contribution is
the composition: we quantify the rule's variance \emph{on no-peak curves},
show that attention structure is stop-invariant while endpoint robustness is not, and build the
matched-accuracy coordinate that the invariant/variant split demands.

\section{Setup}\label{sec:setup}

\paragraph{Registered hypotheses.}
The registration fixed three hypotheses, each a directional prediction: that distilled
attention would be \emph{more} overfocused than weight-transferred attention; that the
layer-wise overfocusing signature would predict the ID-to-OOD gap; and that diversifying the
distilled structure would close the gap. The conditions and attention diagnostics below
instrument the first two; the diversification dose ladder is the third's instrument. They are
adjudicated against their registrations in Section~\ref{sec:faithful},
Appendix~\ref{app:prereg}, and Section~\ref{sec:causal}, respectively.

\begin{figure}[b]
\centering
\includegraphics[width=\textwidth]{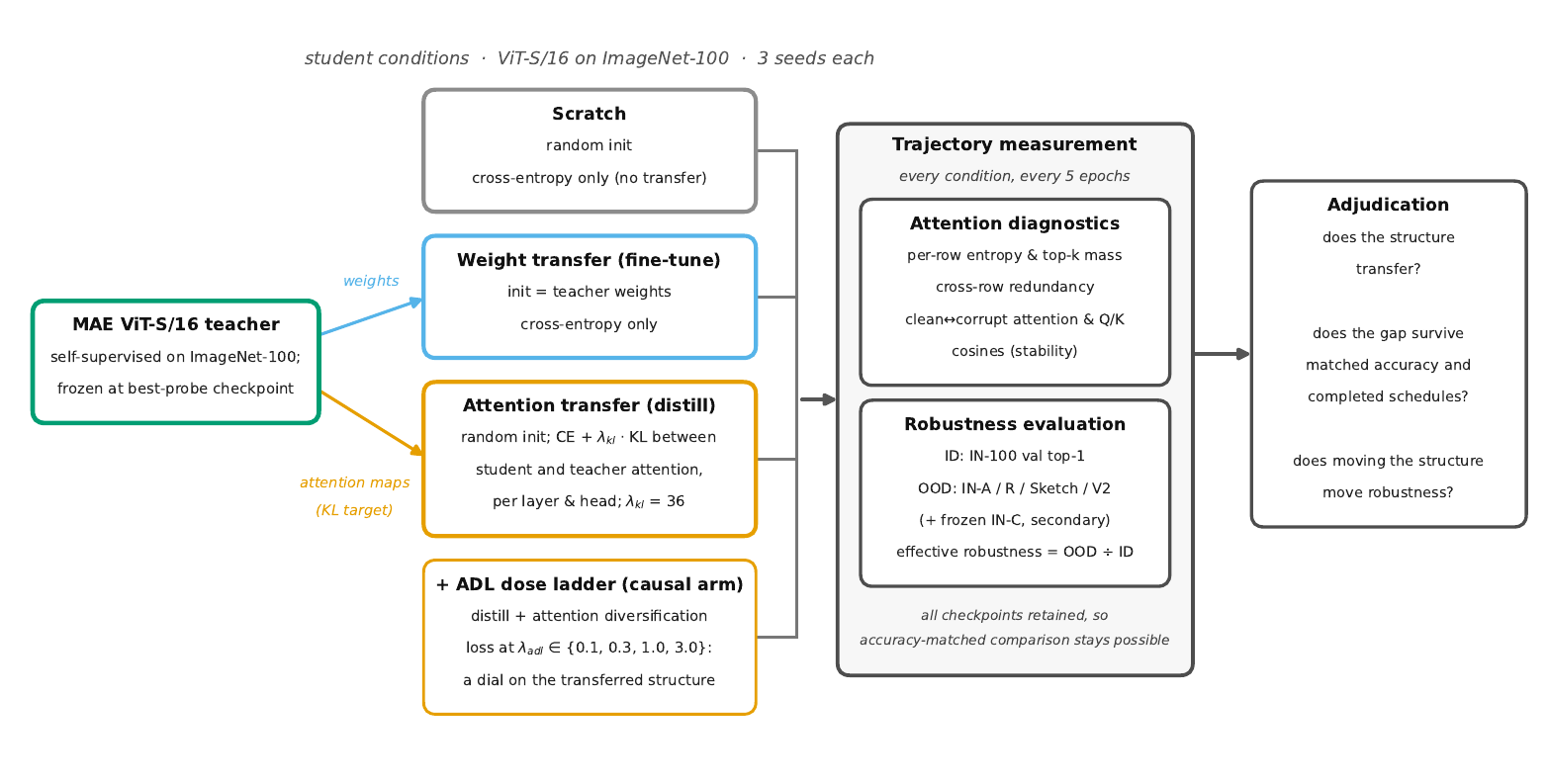}
\caption{Study design: one probe-selected MAE teacher; three training conditions
(scratch, weight transfer, attention transfer) $\times$ three seeds under one
pre-registered stopping rule; an ADL dose ladder on the distillation objective;
full-trajectory checkpoint retention feeding the matched-accuracy analyses; and
labeled full-schedule reruns as the calibration layer.}
\label{fig:pipeline}
\end{figure}

\paragraph{Conditions.}
We compare three ways of training a ViT-S/16 classifier \citep{dosovitskiy2021image, touvron2021training} on
ImageNet-100 (Figure~\ref{fig:pipeline}): training from random
initialization (\emph{scratch}); initializing every weight from a self-supervised teacher and
fine-tuning (\emph{weight transfer}, the ``fine-tune'' condition of \citealt{li2024attention});
and training from random initialization while distilling only the teacher's attention maps
(\emph{attention transfer}, ``distill''). The distillation term is a per-layer, per-head KL
divergence between student and teacher attention distributions, added to cross-entropy at
weight $\lambda_{\mathrm{kl}} = 36$.
This value is not a tuned hyperparameter but the arithmetic equivalent of the official per-layer
strength of \citet{li2024attention}, with $\lambda = 3$ summed over layers and converted to our
mean-over-layers formulation. We verified that our loss reproduces the official implementation's
gradients to float precision, with a maximum element difference of $5.6 \times 10^{-9}$
(Appendix~\ref{app:lambda}). The causal intervention of Section~\ref{sec:causal} augments the
distillation objective with the Attention Diversification Loss (ADL) of \citet{guo2023robustifying}
at doses $\lambda_{\mathrm{adl}} \in \{0.1, 0.3, 1.0, 3.0\}$, applied per-head to the same
attention quantity our diagnostics measure. This matches Guo et al.'s Eq.~1--2; their released
implementation averages heads before the loss, and we keep the per-head form so the penalized
quantity is identical to the measured one. Three seeds per condition throughout.

\paragraph{Teacher.}
The teacher is a ViT-S masked autoencoder \citep{he2022masked}, pretrained from scratch on
ImageNet-100 for up to 400 epochs with linear-probe-based early stopping (probes every 25
epochs from epoch 24, patience 3). The rule promoted the checkpoint with the best in-training probe, epoch
299; three later probes were higher by at most 0.0034, which the rule's 0.005 min-delta
counts as no improvement. Because every downstream run depends on this one checkpoint, a
quality gate guarded the spending against teacher regression or mis-loading: its mechanism
was pre-registered, and its floor was pinned at launch as the teacher's measured probe
accuracy minus a safety margin, 0.48 against a measured 0.5556. The two probe accuracies come
from two instruments: the 0.5556 is a full 50-epoch linear head trained on the promoted
checkpoint, while the in-training probes read by the stopping rule use a shorter head and sit
lower, at 0.5434 for the promoted checkpoint. Both transfer conditions inherit this single
checkpoint, so variation in teacher quality cannot differentiate them; the comparison is made
at this one teacher's quality (Section~\ref{sec:limitations}).

\paragraph{Training protocol.}
All conditions train under one pre-registered early-stopping rule, inside per-condition cosine
horizons. The rule evaluates validation top-1 every 5 epochs with a min-delta of 0.1pp and a
patience of 5 evaluations, that is, 25 epochs of demonstrated flatness. The random-init family
runs at 300 epochs, the standard DeiT-style schedule \citep{touvron2021training}, fixed before
any run; in hindsight that is 2.07$\times$ the first observed distillation stop, though later
seeds stopped as late as epoch 290. Fine-tuning runs at 200: its initial 100-epoch schedule
ran out while validation was still climbing and was doubled once under a pre-declared
pilot-calibration clause, landing at roughly twice its observed plateau epoch. Runs that stop early carry the rule's early-stop certificate; runs
that complete their horizon are reported at their best checkpoint with the cap-hit flagged and the
terminal 25-epoch net gain disclosed (Appendix~\ref{app:convergence}). Convergence claims never
rest on the stopping rule's verdicts: they rest on completed or completion-bounded anneals,
directly shown terminal flatness, seed-band replication, and the measured cost of the
interruption where the rule interrupted a schedule (Section~\ref{sec:gap}). Checkpoints are retained
every 5 epochs, full precision, no pruning: every accuracy level any run passed through remains a
loadable model, which is what makes the matched-accuracy analyses below possible. The full
training configuration, hardware, and compute accounting are in Appendix~\ref{app:config}.

\begin{figure}[t]
\centering
\includegraphics[width=\textwidth]{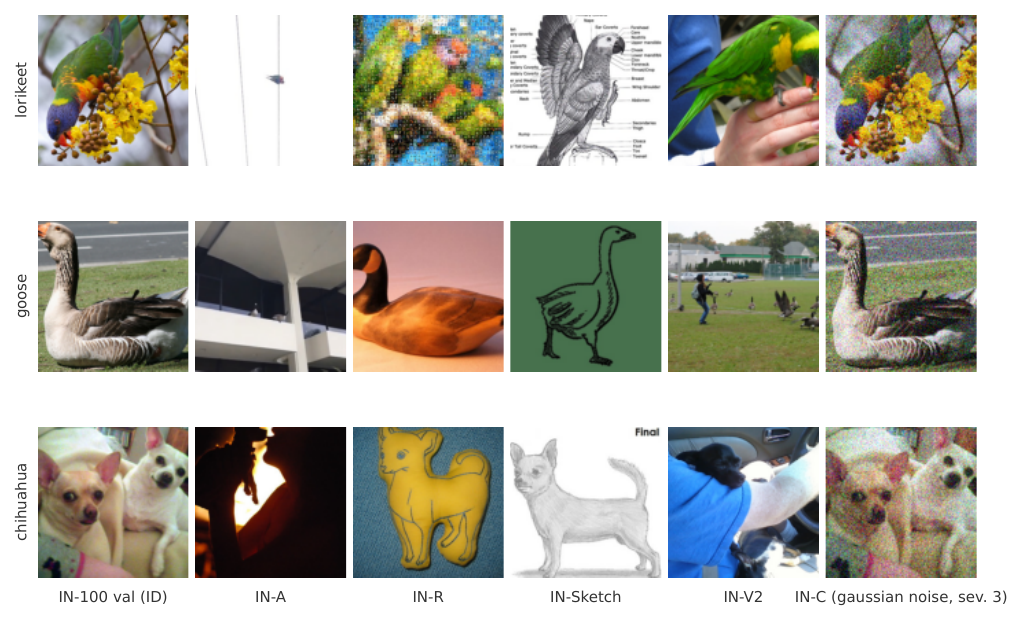}
\caption{Three classes across the evaluation suites: the ID validation image
and the same class as it appears in each natural-shift suite, plus an IN-C
panel rendered from the ID image with the study's reference corruption
implementation (gaussian noise, severity 3). Selection is by fixed rule, with
no per-image curation: the classes are the first three, in sorted WordNet-ID
order, of the IN-100 classes covered by both IN-A and IN-R, and each panel is
the first file in sorted filename order in its class directory, shown under
the evaluation crop.}
\label{fig:ood_examples}
\end{figure}

\paragraph{Evaluation.}
In-distribution (ID) accuracy is top-1 on the held-out ImageNet-100 validation set.
Out-of-distribution (OOD) evaluation uses the natural-shift suite: ImageNet-A (natural
photographs hard enough to defeat a standard classifier), ImageNet-R (renditions of the
classes: art, cartoons, sketches, sculpture), ImageNet-Sketch (sketches and line drawings), and
ImageNet-V2 (a fresh test set collected under the original ImageNet protocol)
\citep{hendrycks2021many, hendrycks2021natural, wang2019learning,
recht2019imagenet}, label-mapped to the covered classes (A/R/Sketch/V2 cover 15/19/100/100
of the 100 classes). Figure~\ref{fig:ood_examples} shows three classes across every suite. We report
\emph{effective robustness}: OOD accuracy divided by ID accuracy on the covered classes; of the
skill a model has, the fraction that survives distribution shift. We use the term for this ratio;
it differs from the residual-above-baseline definition of \citet{taori2020measuring} and
\citet{miller2021accuracy}, and we adopt the ratio for scale-independence within our
checkpoint-matched comparisons and flag the collision explicitly here. Our primary
comparisons are checkpoint-matched or near-equal in ID, where the two definitions order
models identically; where compared models differ in ID (the endpoint coordinate and the
maturity correlation), the definitions can disagree, and Appendix~\ref{app:residual}
reports the residual-form recomputation. Uncertainty on every
endpoint quantity comes from a paired image-level bootstrap over the per-image outcomes of all
evaluated models ($B = 10{,}000$, percentile 95\% intervals, one index draw per replicate applied
to every model, ID resamples propagated into the covered-class denominators). These intervals
quantify image-sampling variability on the fixed benchmarks, conditional on their class
composition; the covered-class sets are properties of the frozen benchmarks and are held fixed. The per-image outcome dump backing the bootstrap covers
the endpoint models only, so matched-coordinate quantities computed on trajectory checkpoints
(Sections~\ref{sec:matched} and~\ref{sec:causal}) carry no bootstrap intervals; their
uncertainty is carried by the single-pair noise scale ($\pm$0.5--1pp) and the disclosed seed
spreads. A
frozen, self-generated ImageNet-C \citep{hendrycks2019benchmarking} provides a secondary
synthetic-shift probe on role-labeled representative checkpoints (reference implementation, PNG,
byte-identical across all conditions and evaluations).

Attention structure is measured on the
validation set with three quantities. Per-row entropy and top-$k$ mass record how sharply each
token focuses. Cross-row redundancy, Guo et al.'s Eq.~1--2 applied per head, records whether
different tokens stare at the same places. Clean-corrupt cosines of attention maps and of the
Q/K projections record whether routing holds still under corruption. All selection decisions (stopping, checkpoint promotion, the teacher, thresholds) use
ID validation only; OOD suites are touched exactly once per evaluated checkpoint.

\paragraph{Comparisons.}
We use two co-primary coordinates, both fixed in the analysis code before any training run.
The \emph{endpoint} coordinate takes each method's rule-governed best, which is the coordinate
closest to the literature's practice, though not identical to it: Li et al.\ compare fixed
budgets with no stopping rule, while ours is rule-governed. It is always reported with the qualifier ``under the
pre-registered rule and horizons.'' The \emph{matched-accuracy} coordinate compares models at
common ID accuracy using trajectory checkpoints, matched in accuracy by construction; the
residual schedule-state sensitivity is measured in Section~\ref{sec:matched}. A calibration
layer of full-schedule reruns prices the stopping rule's interruptions. These
``continuations'' retrain the same seed and schedule from epoch 0 with stopping disabled, with
replay fidelity checked against the canonical prefix. They are labeled artifacts that never
replace canonical numbers: first-class evidence, second-class authority.

\section{Results}\label{sec:results}

\subsection{The transfer is faithful}\label{sec:faithful}

\begin{figure}[b]
\centering
\includegraphics[width=0.9\textwidth]{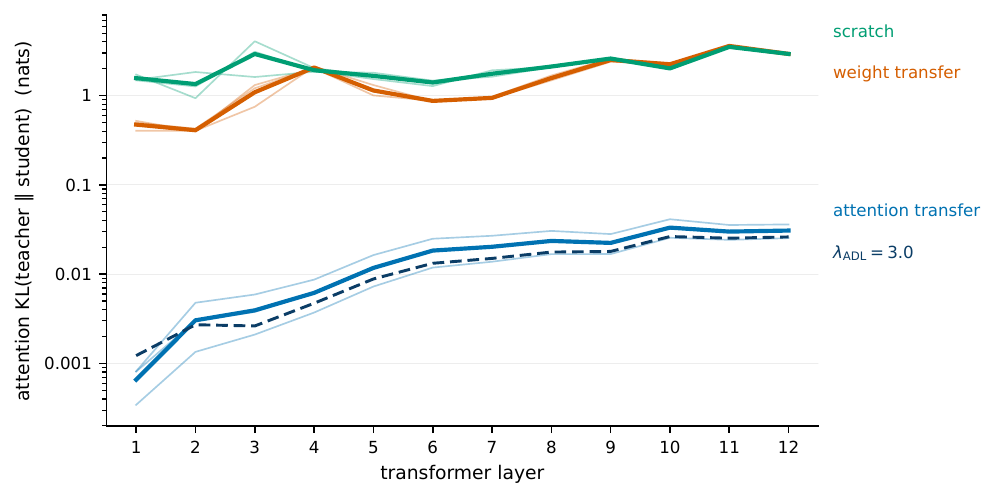}
\caption{Per-layer attention KL(teacher$\Vert$student) on held-out validation
images (log scale; thin lines are seeds, bold lines seed means). The distilled
students sit roughly two orders of magnitude below both contrasts at every
layer; the maximum-dose ADL students (dashed) track the dose-zero distilled band
(mean 0.0135 vs.\ 0.0170 nats).}
\label{fig:fidelity}
\end{figure}

Attention transfer transfers attention, with tight and durable fidelity. The direct
measurement is eval-time divergence on held-out validation images
(Table~\ref{tab:fidelity}): the distilled student's attention sits at a mean per-layer,
per-head KL of 0.017 nats from its teacher's (0.012--0.022 across seeds, map cosine 0.99),
against 1.65 nats for fine-tuning and 2.15 for scratch. The student trained to copy the
teacher's gaze ends up 97$\times$ closer to it than the student that \emph{started from the
teacher's own weights}: the transplant held while the inheritance wandered. Per-layer attention KL
never exceeds 0.042 nats in any layer of any seed
(Figure~\ref{fig:fidelity}), so the mean is not hiding a broken layer. Every other structural
measure agrees. Mean cross-row redundancy lands within 0.002 of the teacher's
value (0.468 against 0.4665, about 1/25 of fine-tuning's drift, worst seed
+0.0026), and per-row entropy and top-$k$ mass are likewise pinned: 3.61 against the teacher's
3.60, and 0.415 against 0.418. Corruption-stability \emph{exceeds} every other student's
(Table~\ref{tab:structure}). The distillation term itself
documents the mechanism: the per-row KL collapses from 1.67 at initialization to 0.03 by
epoch 40 and sits near 0.02 for the rest of training (0.022 at the stop). What
little divergence remains is not spread uniformly: it concentrates in the CLS rows of the deepest
layers, the rows pressured by the student's classification readout. The student wears the teacher's
gaze everywhere and spends its small KL slack sharpening the classifier's readout rows.

This faithfulness is the study's fixed point. It replicates across all three seeds, and it is
invariant to training duration: students whose stopping epochs ranged from 145 to 290, and
whose accuracies span 5.6 percentage points, carry the same attention structure to within
measurement noise, with redundancy varying by $\pm 0.0008$ and entropy by $\pm 0.006$ across
seeds. And it tightens slightly with training, with per-seed KL 0.022/0.017/0.012, ordered by
stop epoch. Structure locks in early
and permanently; everything else about the model keeps maturing around it
(Section~\ref{sec:gap}). The explanation for the robustness gap that follows cannot be a
failure to copy: the copy is verified, stable, and, by the standards the literature uses to
score attention quality, the best-looking attention in the family.

\begin{figure}[t]
\centering
\includegraphics[width=\textwidth]{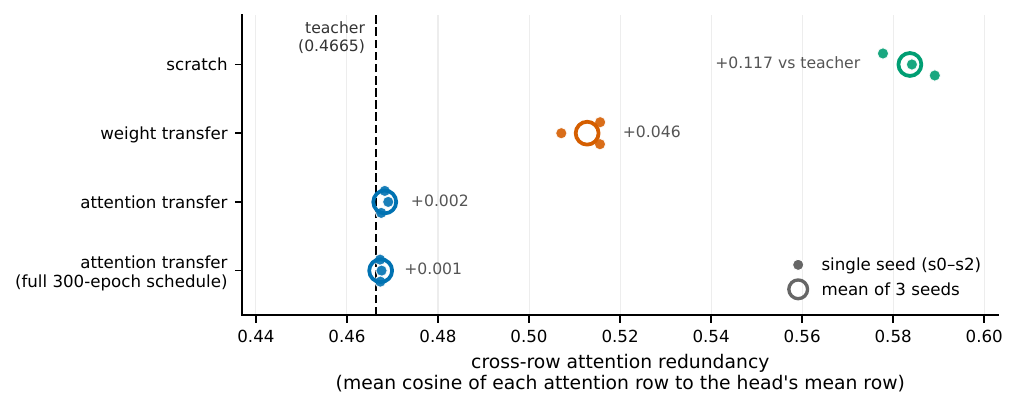}
\caption{Cross-row redundancy at rule-governed endpoints (mean cosine of each
thresholded attention row to the head's mean row, Guo Eq.~1--2; dots: seeds; open
markers: seed means; dashed line: teacher). Distilled students sit at the
teacher's value, including after full 300-epoch schedules; supervised training
drifts redundancy upward when unconstrained.}
\label{fig:ladder}
\end{figure}

Two corollaries frame everything downstream. First, the two senses of ``overfocusing''
dissociate: transferred attention is the \emph{most} concentrated per-row (entropy 3.61 vs.\
fine-tune's 4.14 and scratch's 4.34) and the \emph{least} redundant across rows (0.468 vs.\
$\sim$0.51 and $\sim$0.58). We had registered the prediction that transferred attention would
be more overfocused than weight-transferred attention. Per row it is; across rows it is the
reverse. The prediction was not answerable as posed, which is the finding. Sharpness and redundancy are different properties; they are anti-associated across our
conditions (the sharpest attention is the least redundant, teacher and distilled student
both) and, as Section~\ref{sec:causal} shows, independently manipulable. Second,
supervised training \emph{increases} redundancy when left free: fine-tuning drifts from the
teacher's 0.4665 to $\sim$0.51 over its schedule; scratch, never anchored, reaches $\sim$0.58
(Figure~\ref{fig:ladder}). The ``textbook-diverse'' attention the literature admires is what
reconstruction pretraining builds and what classification, unconstrained, erodes.

\begin{table}[b]
\centering
\caption{Eval-time transplant fidelity on held-out validation images: mean per-layer/head KL(teacher$\Vert$student) and cosine similarity of attention maps. Mean (min--max) over seeds.}
\label{tab:fidelity}
\begin{tabular}{lllr}
\toprule
Student & KL(t$\Vert$s) (nats) & Cosine & $n$ \\
\midrule
Attention transfer & 0.0170 (0.0124--0.0217) & 0.989 (0.986--0.992) & 3 \\
~~+ ADL $\lambda{=}3.0$ & 0.0135 (0.0126--0.0141) & 0.991 (0.990--0.991) & 3 \\
Weight transfer & 1.6545 (1.6243--1.6800) & 0.499 (0.494--0.507) & 3 \\
Scratch & 2.1482 (2.0881--2.2270) & 0.342 (0.339--0.345) & 3 \\
\bottomrule
\end{tabular}
\end{table}

\begin{table}[b]
\centering
\caption{Attention structure at rule-governed endpoints (seed means; per-seed dispersion in Appendix~\ref{app:lottery}). Entropy and top-$k$ mass measure per-row concentration; cross-row redundancy is Guo et al.'s diversity quantity (higher = rows more alike); stability is the clean--corrupt attention cosine.}
\label{tab:structure}
\begin{tabular}{lrrrr}
\toprule
Model & Entropy & Top-$k$ & Redundancy & Stability \\
\midrule
Teacher (MAE) & 3.596 & 0.418 & 0.4665 & -- \\
Attention transfer & 3.610 & 0.415 & 0.4683 & 0.810 \\
Weight transfer & 4.136 & 0.305 & 0.5128 & 0.742 \\
Scratch & 4.338 & 0.257 & 0.5837 & 0.750 \\
\bottomrule
\end{tabular}
\end{table}

\begin{figure}[t]
\centering
\includegraphics[width=0.85\textwidth]{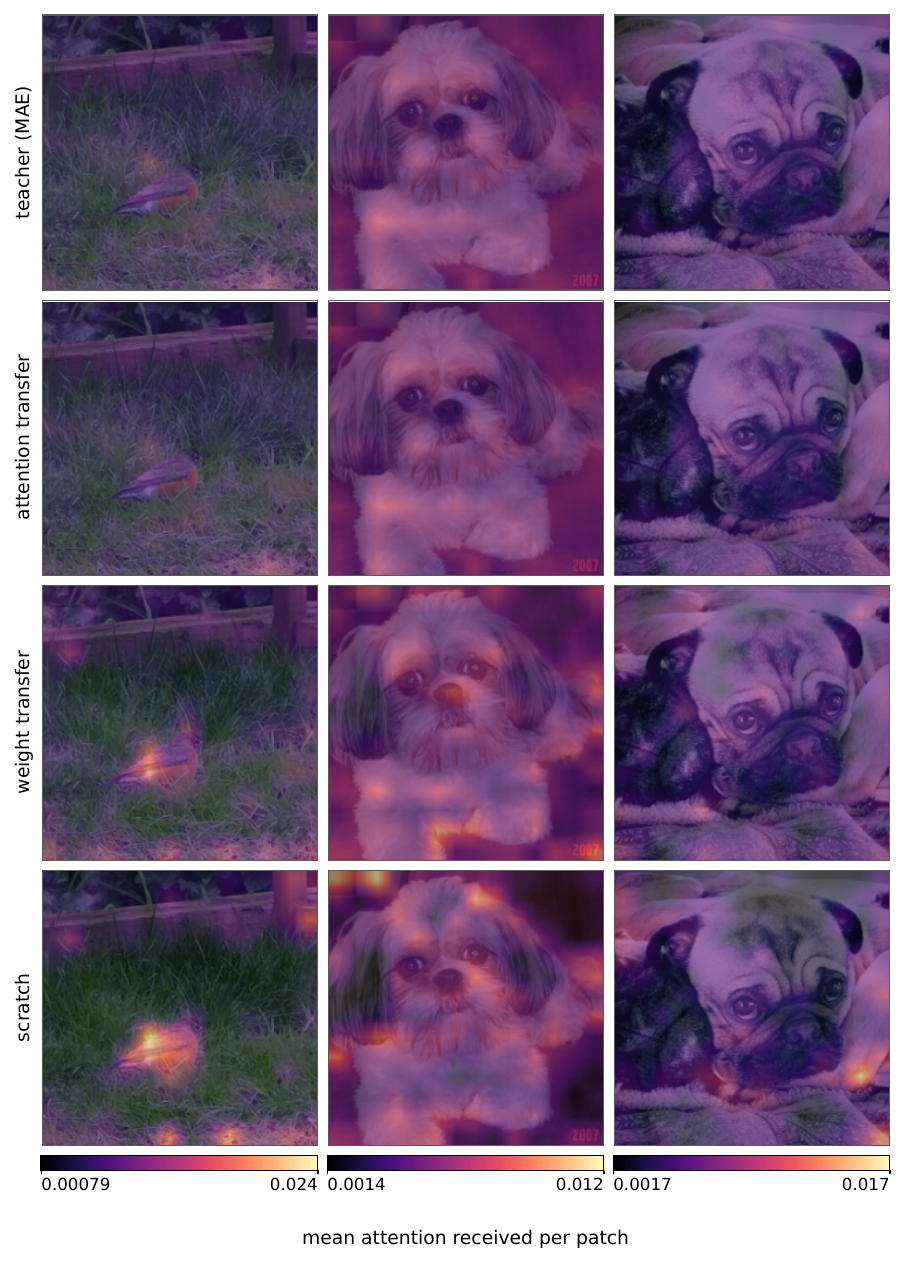}
\caption{Mean attention received per patch (layer 6 of 12, averaged over heads
and patch queries; CLS excluded as query), for three validation images: the
first three of the study's fixed eight-image attention snapshot, with no
per-image selection. The
color scale is shared within each column, so panels are comparable across
models; per-column ranges are printed below. The layer was selected by a fixed
rule (the layer whose teacher--distill similarity is the median across layers),
not for visual effect. The comparison is one of \emph{similarity} only: the
attention-transfer row reproduces the teacher's maps (cosine 0.998--0.999 on
these images); weight transfer and scratch differ from both. Nothing in this
figure bears on which attention is \emph{better}; Section~\ref{sec:causal}
addresses that question causally.}
\label{fig:overlays}
\end{figure}

\subsection{The gap is real, and it has a time axis}\label{sec:gap}

The robustness gap reproduces. At rule-governed endpoints, attention-transfer students trail
weight-transfer students in effective robustness at every seed: gaps of 0.0575 [0.0445, 0.0703],
0.0323 [0.0209, 0.0442], and 0.0181 [0.0073, 0.0289] against a pre-registered threshold of 0.01
(Table~\ref{tab:gaps}). The intervals are paired image-level bootstraps; every one excludes zero, and
every point estimate clears the threshold, though s2's interval reaches below it. Endpoint gaps
are reported here and throughout under the pre-registered rule and horizons. This reproduces the
ordering of \citet{li2024attention} with a 14$\times$ smaller student on a 10$\times$ smaller
dataset: weight transfer above attention transfer above scratch, at every
seed (Table~\ref{tab:endpoints}).

\begin{figure}[t]
\centering
\includegraphics[width=\textwidth]{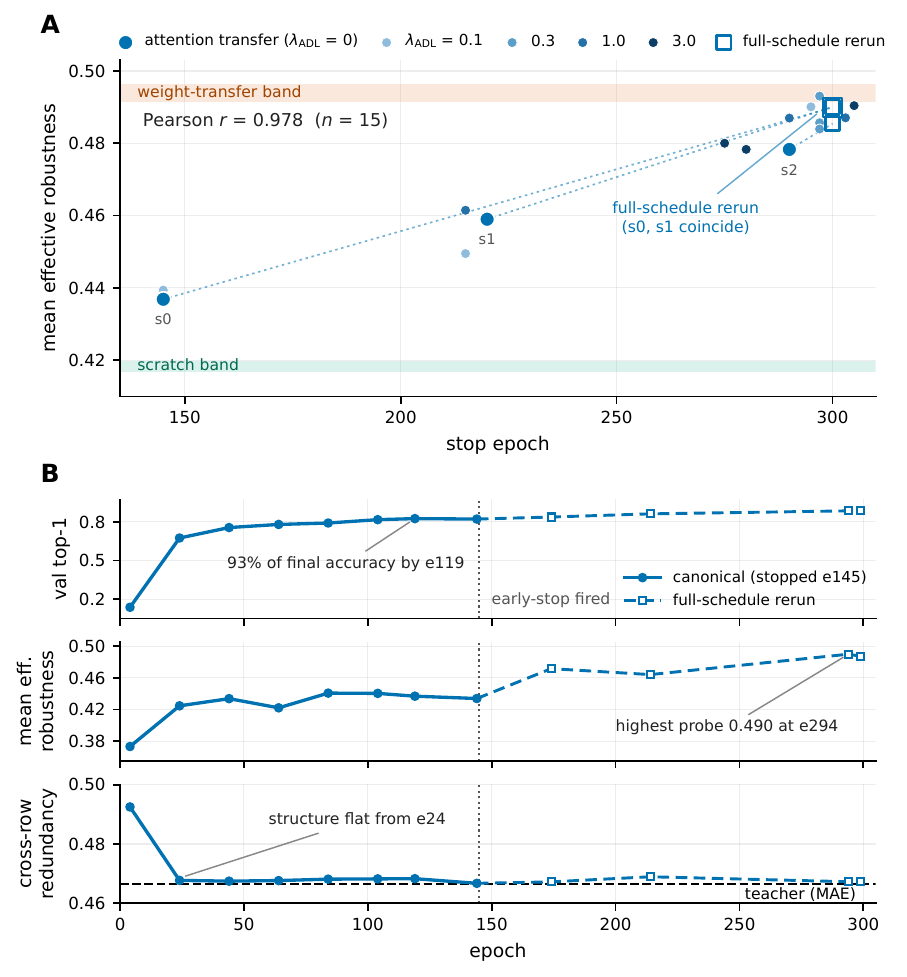}
\caption{\textbf{A:} endpoint effective robustness against stop epoch for all 15
distill-family runs, with $r = 0.978$ computed on true stop epochs. Bands are
min--max of the three weight-transfer and scratch seeds. Open squares are the
full-schedule reruns at epoch 300, with dotted lines connecting each to its same-seed
canonical endpoint; the s0 and s1 reruns coincide to $10^{-4}$ and are drawn
nested. For legibility, the six dosed runs that stop exactly at the 300-epoch
cap are horizontally dodged by $\le$5 epochs; all true stops are exactly 300.
\textbf{B:} three clocks within seed 0, canonical solid and rerun dashed, where the
rerun replays the canonical prefix exactly on shared epochs. Accuracy reaches
93\% of its final value by epoch 119. Effective robustness's highest probe arrives at epoch
294, at 0.490, with the final probe reading 0.487. Cross-row redundancy is flat from epoch 24
at the teacher's value. The dotted vertical line marks the rule's cut at epoch 145.}
\label{fig:maturity}
\end{figure}

But the per-seed gaps are not three estimates of one number. They are three samples of a
\emph{curve}. The distilled runs' stopping epochs ranged from 145 to 290, set by an early-stopping rule whose
variance we characterize in Appendix~\ref{app:lottery}. Endpoint robustness tracks stopping
epoch with $r = 0.978$ across all fifteen distill-family runs
(Figure~\ref{fig:maturity}A), 3 dose-zero and 12 dosed, using
ratio-form effective robustness throughout. The relation is not an artifact of
the pooling: the twelve $\lambda_{\mathrm{adl}} \le 1.0$ runs alone give 0.977, the
dose-zero-plus-$\lambda$0.1 restriction gives 0.973, and the nine rule-stopped runs, with
all six horizon-capped runs dropped, give 0.974.
Later-stopping runs are more robust. The endpoint gap versus fine-tuning shrinks monotonically
as distillation approaches schedule completion, 0.057 $\to$ 0.032 $\to$ 0.018, while the
attention structure never moves (Section~\ref{sec:faithful}). Whatever matures is not the
routing.

\paragraph{Three clocks.}
A densified evaluation pass separates three timescales inside one run. It covers one seed's
saved trajectory checkpoints, its canonical run plus its labeled full-schedule rerun
(Figure~\ref{fig:maturity}B). Attention
structure locks first: redundancy reaches its final value by epoch 24 and stays within
$\pm 0.002$ through epoch 299 (full post-epoch-24 range 0.4667--0.4690, 5\% of the
between-condition separation). Accuracy matures next: 93\% of final ID accuracy by epoch 119.
Robustness matures last: its highest probe arrives at epoch 294 of 300 (0.490; the final
probe reads 0.487, within per-probe noise). In the 55-epoch window bracketing the rule's cut (epochs 119--174, from the promoted
checkpoint to 29 epochs past the rule's firing), the same seed gained 3.5pp of effective robustness
against 1.2pp of ID accuracy. Over their full remaining schedules, measured by their labeled
reruns, the two later seeds are same-signed: +3.1 against +1.8pp, and +0.7 against +0.3pp
(Appendix~\ref{app:continuation}).
In this regime, robustness is the slowest-maturing quantity in the model, and an early-stopping
rule tuned to accuracy flatness systematically undersamples it.

\paragraph{Completing the schedules.}
The maturity reading predicts that the gap at \emph{completed} schedules should be far smaller
than the endpoint gap. We tested this within-seed: for each distilled seed, a full-schedule
rerun (same seed, same schedule, stopping disabled) extends the canonical run to its 300-epoch
horizon. Replay fidelity was exact: across 131 shared evaluations, the reruns reproduce their
canonical prefixes to all reported decimals, so each rerun \emph{is} the canonical run
continued. The interruption cost is a dose-response in how hot the rule cut. The seed cut at
59\% of peak learning rate recovered +6.0pp ID and +5.3pp effective robustness over its
remaining 155 epochs. The seed cut at 20\% recovered +1.8 and +3.1pp. The seed cut at
1.3\% of peak moved +0.3 and +0.7pp; that seed is an in-design negative control, cut 10 epochs
from the horizon at the schedule's 1\% floor plus 0.3\% of the remaining decay. Robustness
outgained accuracy in both later seeds' remainders and in seed 0's post-cut window (+3.5
against +1.2pp); only over seed 0's full 155-epoch remainder did the two move together, at
+5.3 against +6.0pp.
Meanwhile the transferred structure stayed immobile through all three full schedules
(redundancy 0.4673/0.4676/0.4673 at the completed schedules' promoted checkpoints, still
at the teacher's 0.4665).

At completed schedules the gap closes to below the pre-registered threshold in two of three
seeds: 0.0044
[$-$0.0059, +0.0148], 0.0013 [$-$0.0096, +0.0122], and 0.0110 [+0.0003, +0.0217] per seed
(Table~\ref{tab:gaps}). The first two sit below the 0.01 threshold and are statistically
indistinguishable from zero at image-level noise; the third sits at 1.1$\times$ the threshold,
and pairs the strongest fine-tuned seed (0.4964) with the weakest completed distilled one. The
completed-schedule bands are adjacent: distill 0.4854--0.4901 vs.\ fine-tune 0.4913--0.4964,
about 0.55pp between means. The endpoint-coordinate triple (0.0575/0.0323/0.0181) and the
completed-schedule triple (0.0044/0.0013/0.0110) are both real and are never blended: the first
is what a practitioner's stopping rule sees, the second is what the methods produce at equal,
completed budgets. The conclusion this licenses: \textbf{the endpoint robustness gap at this
scale is substantially a training-maturity artifact}. Under the pre-registered rule and
horizons the gap is real and in line with Li et al.'s; at completed schedules it compresses below our
registered threshold in two of three seeds, and comparing the methods at equal accuracy gives
the same result (Section~\ref{sec:matched}). For seed 0, 92\% of the endpoint-coordinate gap (0.0575 $\to$
0.0044) was maturity, not method.

\begin{table}[b]
\centering
\caption{Rule-governed endpoints per condition and seed, with labeled full-schedule reruns. Effective robustness is the 4-suite mean; brackets are paired image-level bootstrap 95\% CIs (B=10{,}000). The evaluated checkpoint is each run's min-delta-promoted best, which may precede the stop epoch shown.}
\label{tab:endpoints}
\begin{tabular}{llrrl}
\toprule
Condition & Seed & Stop epoch & ID top-1 & Eff.\ robustness [95\% CI] \\
\midrule
Scratch & s0 & 300 (cap) & 0.8498 & 0.4199 [0.4070, 0.4332] \\
 & s1 & 300 (cap) & 0.8496 & 0.4168 [0.4038, 0.4301] \\
 & s2 & 290 (early stop) & 0.8462 & 0.4188 [0.4053, 0.4324] \\
\addlinespace
Weight transfer (ft) & s0 & 200 (cap) & 0.8916 & 0.4943 [0.4814, 0.5073] \\
 & s1 & 200 (early stop) & 0.8896 & 0.4913 [0.4785, 0.5044] \\
 & s2 & 200 (cap) & 0.8926 & 0.4964 [0.4836, 0.5092] \\
\addlinespace
Attention transfer (distill) & s0 & 145 (early stop) & 0.8248 & 0.4369 [0.4233, 0.4510] \\
 & s1 & 220 (early stop) & 0.8610 & 0.4590 [0.4461, 0.4720] \\
 & s2 & 290 (early stop) & 0.8804 & 0.4783 [0.4658, 0.4910] \\
\addlinespace
~~distill, full-schedule rerun & s0 & 300 (no rule) & 0.8850 & 0.4900 [0.4773, 0.5023] \\
 & s1 & 300 (no rule) & 0.8786 & 0.4901 [0.4774, 0.5030] \\
 & s2 & 300 (no rule) & 0.8832 & 0.4854 [0.4727, 0.4980] \\
\bottomrule
\end{tabular}
\end{table}

\begin{table}[t]
\centering
\caption{The robustness gap (ft $-$ distill, mean effective robustness) per seed, in the endpoint coordinate (rule-governed stops) and at completed schedules (full-schedule reruns). Pre-registered threshold 0.01. Paired bootstrap 95\% CIs.}
\label{tab:gaps}
\begin{tabular}{lll}
\toprule
Seed & Endpoint gap & Completed-schedule gap \\
\midrule
s0 & +0.0575 [0.0445, 0.0703] & +0.0044 [-0.0059, 0.0148] \\
s1 & +0.0323 [0.0209, 0.0442] & +0.0013 [-0.0096, 0.0122] \\
s2 & +0.0181 [0.0073, 0.0289] & +0.0110 [0.0003, 0.0217] \\
\bottomrule
\end{tabular}
\end{table}

\subsection{Matched-skill comparisons}\label{sec:matched}

\begin{figure}[t]
\centering
\includegraphics[width=0.75\textwidth]{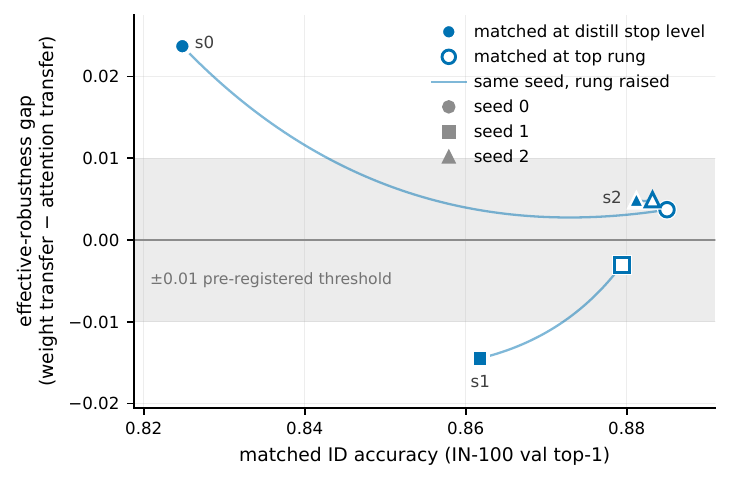}
\caption{The matched-accuracy gap per seed at its own matched level (filled)
and at the top matched level (open); connectors join each seed's two measured rungs and are
annotations, not measured paths. Shaded band = pre-registered $\pm$0.01 threshold. Gaps of both signs converge into the band
as the matched level rises.}
\label{fig:level_ladder}
\end{figure}

The endpoint coordinate compares models that reached different accuracies; the
matched-accuracy coordinate removes that. Using the retained trajectory checkpoints, we compare fine-tuned and distilled models of the
same seed at common ID accuracy, at each seed's own matched level and at the top level both
trajectories reach (Table~\ref{tab:ladder}; Figure~\ref{fig:level_ladder}). On the distilled
side the top rungs are reached only by the labeled full-schedule reruns, which supply those
checkpoints.

At low matched levels the per-seed gaps are large \emph{and of both signs}: +0.0237 at 0.8248
(s0), $-$0.0145 at 0.8618 (s1), +0.0049 at 0.8812 (s2). The s1 rung is the study's one loose
match: its fine-tuned side sits 0.0054 below target against the 0.005 selection tolerance,
which inflates that negative gap by roughly 0.6pp on a linear read of the trajectory
(Appendix~\ref{app:suites}). The spread is not dose or method but robustness-arrival
heterogeneity in the baselines themselves: at the common level 0.8248 the three distilled
seeds' effective robustness spans 0.437--0.480, a 4.3pp seed range. The three same-level
gaps built from it are +2.4, $-$0.9, and $-$3.9pp, a spread of $\pm$3.2pp, far beyond single-pair
measurement noise of $\pm$0.5--1pp. A single-level matched read is therefore seed-noisy by construction
(the registered analysis at this level labels its own mean of $-$0.8pp a candidate null), and we
decline to summarize it with one number.

Maturity resolves the heterogeneity. At the top matched rungs ($\sim$0.88, the levels only
completed schedules reach) the gaps are +0.0037, $-$0.0030, +0.0049: mean +0.002, all inside
the $\pm$0.01 threshold band, still mixed in sign. The one negative sign rests on a
registered selection detail. The s1 top rung selects the rerun's epoch-299 checkpoint, at ID
accuracy 0.8794 and effective robustness 0.4947, rather than its min-delta-promoted best, at
0.8786 and 0.4901. With the promoted endpoint instead, the s1 gap reads +0.0016, and all three
top-rung gaps stay inside the band under either choice. Within-seed, the matched gap converges
toward zero from both directions as level rises: s0 from +2.4pp to +0.4pp, s1 from $-$1.5pp to
$-$0.3pp (s2's two rungs nearly coincide in level, 0.8812 and 0.8832, and read +0.5pp at
both). Skill-matching at mature levels agrees with the completed-schedule endpoint read: the
remaining differences sit inside the pre-registered threshold band (s2's completed-schedule
interval marginally excludes zero, Table~\ref{tab:gaps}).

One honesty bound accompanies the top rungs: the fine-tuned checkpoints there sit at
$\sim$75--82\% of their annealing schedule rather than at completion. We measured this
schedule-state effect on artifacts that already existed, comparing the fine-tuned seed-0
run's epoch-129 checkpoint, mid-anneal at 0.8710, against the completed, fully-annealed
endpoint of the retired 100-epoch horizon, 0.8718, which the archive retains
(Appendix~\ref{app:convergence}). The difference is 2.0pp of effective robustness at
matched ID. The top-rung
checkpoints sit far closer to completion, so we estimate, rather than measure, the residual
effect there at $\lesssim$1pp, of the same order as the top-rung gaps themselves. In the one case measured, the less-complete schedule scored higher, so the
schedule-state bias inflates rather than hides these gaps, making them conservative. But the ladder cannot
resolve differences below this band, which is why the completed-schedule endpoint read, where
both sides are fully annealed, carries the primary weight.

\begin{table}[b]
\centering
\caption{The matched-accuracy ladder (wave-2): gap (ft $-$ distill, mean effective robustness) at each seed's own matched level and at the top matched level, trajectory checkpoints both sides; top-rung distill checkpoints come from the labeled full-schedule reruns. Selection is by nearest ID to the target level; for s1's top rung this picks the rerun's epoch-299 checkpoint rather than its promoted best, and the gap stays inside the band under either choice (Section~\ref{sec:matched}). Bootstrap intervals do not exist for trajectory checkpoints (Section~\ref{sec:setup}).}
\label{tab:ladder}
\begin{tabular}{lrr|rr}
\toprule
& \multicolumn{2}{c|}{Own-level rung} & \multicolumn{2}{c}{Top rung} \\
Seed & level & gap & level & gap \\
\midrule
s0 & 0.8248 & +0.0237 & 0.8850 & +0.0037 \\
s1 & 0.8618 & -0.0145 & 0.8794 & -0.0030 \\
s2 & 0.8812 & +0.0049 & 0.8832 & +0.0049 \\
\bottomrule
\end{tabular}
\end{table}

\subsection{The causal test}\label{sec:causal}

\begin{figure}[t]
\centering
\includegraphics[width=\textwidth]{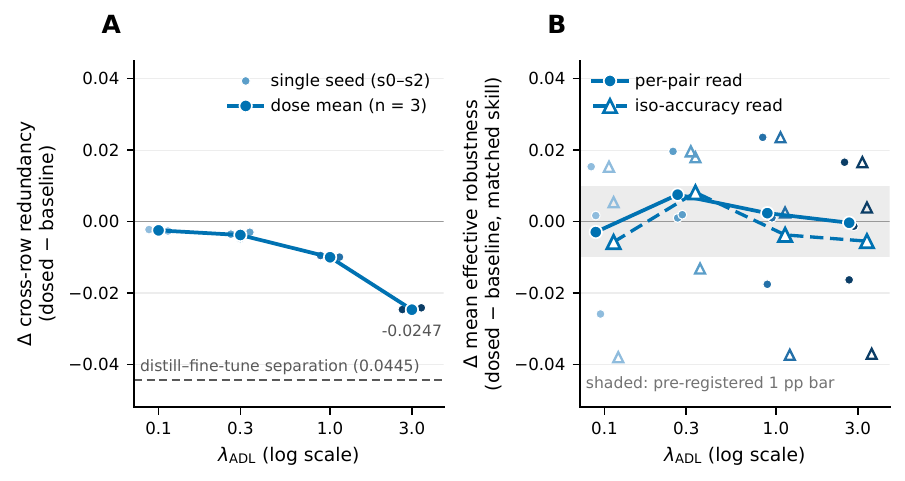}
\caption{The causal test. \textbf{A:} paired same-seed change in cross-row
redundancy per dose, with points for seeds and the line for dose means, and dot shade tracking
$\lambda$. The top dose displaces structure by half the distill--fine-tune separation. The
dashed line is a magnitude ruler only: fine-tuning's redundancy lies in the opposite direction,
at $+0.0445$.
\textbf{B:} matched-skill change in mean effective robustness for the same
runs, per-pair (solid) and iso-accuracy (dashed) reads, on the same vertical
scale as A. The shaded band is the pre-registered 1pp bar. The dial sweeps half
a condition-width of structure, and robustness does not follow.}
\label{fig:dose}
\end{figure}

Everything so far is observational: distilled attention structure and the robustness deficit
co-occur. The intervention asks whether the structure \emph{causes} any of it. Our registered
prediction was that it does: that diversifying the transferred structure would narrow the
gap. Adding Guo
et al.'s diversification penalty to the distillation objective at doses
$\lambda_{\mathrm{adl}} \in \{0.1, 0.3, 1.0, 3.0\}$ turns the one visible knob: cross-row
redundancy (Table~\ref{tab:dose}; Figure~\ref{fig:dose}).

\paragraph{The dial turns.}
Paired same-seed displacement of cross-row redundancy is monotone in dose and accelerating:
$-$0.0025, $-$0.0038, $-$0.0100, $-$0.0247 (dose means, $n{=}3$ each, per-seed values within
0.0011 of their dose mean). The top dose moves redundancy by half the entire
distill--fine-tune condition separation (0.0247 of 0.0445), and it sets the
structure near-deterministically: the three $\lambda{=}3.0$ students land at redundancy
0.4437/0.4438/0.4434, a $\pm$0.0002 replication across seeds. The instrument is
one-directional: ADL diversifies, pushing redundancy \emph{away} from fine-tuning's higher
value and below the teacher's; the separation fraction is a magnitude yardstick, not a
statement that the student moved toward fine-tuning's structure.

\paragraph{The dial is a scalpel.}
To first order, nothing else moves. Per-row entropy and top-$k$ mass stay family-flat at every
dose, with a maximum entropy shift of 0.015 against a 0.53 condition separation. Eval-time
transplant fidelity is unimpaired at maximum dose, KL 0.0135 against the dose-zero trio's
0.0170; and because fidelity tightens with training and the dosed runs stopped later, the
conservative comparison is the nearest-matched pair, dose-zero at epoch 290 (KL 0.0124)
against $\lambda_{\mathrm{adl}}{=}3.0$ at epoch 300 (KL 0.0126), which shows no fidelity cost
either. ADL reshapes cross-row geometry while per-row inheritance stays intact. ID accuracy is
likewise uncosted in the completed-schedule coordinate at every dose: the dosed runs that reached epoch
300 (three at $\lambda_{\mathrm{adl}}{=}0.3$, one at each of the other doses) land at
0.8822--0.8858, against 0.8786--0.8850 for the labeled dose-zero full-schedule
reruns. Raw endpoint deltas include
$-$2.1 and $-$2.3pp readings, but those are stop-timing artifacts rather than dose effects. One
second-order co-movement is disclosed: at the top dose, corruption-stability shifts by
$-$0.0043 $\pm$ 0.0020, with 3/3 seeds negative, about 6\% of the condition separation, while at
$\lambda_{\mathrm{adl}}{=}0.1$ the mean shift is +0.0005. Even that small shift is partly
timing rather than treatment, since stability drifts by
$\sim$0.005 within a seed over epochs 119--300 and the dosed runs stopped later than their
references. Either way it is an order of
magnitude smaller than the redundancy displacement.

\paragraph{Robustness does not follow.}
Table~\ref{tab:verdict1} is the pre-registered readout: matched-skill $\Delta$ effective
robustness (dosed minus dose-zero, same seed) in the two registered coordinates, per-pair
(each pair at its own matched level) and iso-level (all models at $L^{*} = 0.8248$). Every dose mean
is inside $\pm$1pp in both coordinates: per-pair $-$0.29/+0.75/+0.23/$-$0.03pp, and
iso-level $-$0.57/+0.82/$-$0.37/$-$0.55pp. There is no monotone trend in either read, and
per-seed directions split at $\lambda_{\mathrm{adl}} \ge 1.0$. The two coordinates agree at the
dose-mean level, differing by at most 0.61pp. That agreement, though, does not rest on three
independent contrasts: seed 0's own matched level coincides with $L^{*}$, so its two columns are the same
measurement, while seeds 1 and 2 differ between coordinates by up to 2.0pp and 3.8pp in opposite
directions, discrepancies that cancel in the dose means. Seed 2's iso column is set throughout by one
early-arriving baseline checkpoint at epoch 134, ID accuracy 0.8214 and effective robustness
0.480, which is the previous subsection's arrival heterogeneity expressed as a coordinate
offset. Realized pair scatter ($\pm$1.7--3.8pp) is dominated by the same robustness-arrival
heterogeneity as Section~\ref{sec:matched}, which is why the verdict rests on dose means and
two-coordinate agreement rather than any single pair. Dose means are over three seeds, and their
seed-level 95\% intervals span $\pm$2.6--7.7pp against the 1pp bar. Wide as those intervals
are, a systematic dose effect would have to move the dose means together across the ladder,
and they are flat; the design therefore excludes a monotone effect at the 1pp scale, but not
seed-scale effects of a few pp. That is why the verdict is stated as no detectable response rather than a bounded
effect size.

Read against the verdict definitions frozen before these evaluations ran: \textbf{flat}. Forcing cross-row
redundancy through half of the condition separation produced no detectable robustness response
at matched skill, under either registered way of matching accuracy. This is consistent with the
deficit residing in features, not in the visible routing. It does not prove the features
account, since elimination plus intervention is not direct feature measurement, and its scope
is this regime. But the one component of the transfer everyone can see, the component the
intervention can steer by half a condition-width, produced no detectable causal effect on
robustness here, in the one direction the instrument can push.

\begin{table}[b]
\centering
\caption{The causal dial: paired same-seed change in cross-row redundancy at each ADL dose (sweep endpoint minus its dose-zero baseline). Negative = less redundant.}
\label{tab:dose}
\begin{tabular}{lrrrr}
\toprule
Dose & s0 & s1 & s2 & Mean $\pm$ SD \\
\midrule
$\lambda_{\mathrm{adl}}=0.1$ & -0.0022 & -0.0025 & -0.0027 & -0.0025 $\pm$ 0.0002 \\
$\lambda_{\mathrm{adl}}=0.3$ & -0.0035 & -0.0048 & -0.0030 & -0.0038 $\pm$ 0.0009 \\
$\lambda_{\mathrm{adl}}=1.0$ & -0.0095 & -0.0105 & -0.0100 & -0.0100 $\pm$ 0.0005 \\
$\lambda_{\mathrm{adl}}=3.0$ & -0.0246 & -0.0252 & -0.0241 & -0.0247 $\pm$ 0.0006 \\
\bottomrule
\end{tabular}
\end{table}

\begin{table}[t]
\centering
\caption{The causal test: matched-skill $\Delta$ effective robustness (dosed $-$ dose-zero, same seed) at every ADL dose, in both registered coordinates: per-pair (each pair at its own matched level) and iso-level (all models at $L^{*}=0.8248$). Read against the frozen verdict definitions: flat. Seed 0's own matched level coincides with $L^{*}$, so its two blocks repeat one measurement by construction. Bootstrap intervals do not exist for trajectory checkpoints (Section~\ref{sec:setup}); seed-level 95\% intervals on the dose means span $\pm$2.6--7.7pp.}
\label{tab:verdict1}
\begin{tabular}{lrrrr|rrrr}
\toprule
& \multicolumn{4}{c|}{Per-pair} & \multicolumn{4}{c}{Iso-level} \\
Dose & s0 & s1 & s2 & mean & s0 & s1 & s2 & mean \\
\midrule
$\lambda_{\mathrm{adl}}=0.1$ & +0.0154 & +0.0017 & -0.0259 & -0.0029 & +0.0154 & +0.0054 & -0.0380 & -0.0057 \\
$\lambda_{\mathrm{adl}}=0.3$ & +0.0196 & +0.0010 & +0.0019 & +0.0075 & +0.0196 & +0.0180 & -0.0132 & +0.0082 \\
$\lambda_{\mathrm{adl}}=1.0$ & +0.0236 & -0.0176 & +0.0010 & +0.0023 & +0.0236 & +0.0025 & -0.0373 & -0.0037 \\
$\lambda_{\mathrm{adl}}=3.0$ & +0.0166 & -0.0163 & -0.0013 & -0.0003 & +0.0166 & +0.0039 & -0.0370 & -0.0055 \\
\bottomrule
\end{tabular}
\end{table}

\subsection{Secondary probes}\label{sec:secondary}

\paragraph{Synthetic corruptions.}
ImageNet-C is our secondary, role-labeled probe (Table~\ref{tab:incc}; frozen corruption set,
identical bytes for every evaluation). It reproduces the maturity signature within-seed: the
seed-0 distilled endpoint scores 0.6453 at its epoch-145 cut and 0.7067 after completing the
schedule, a +6.1pp movement with zero structural change. The three completed distilled seeds
cluster at 0.7067--0.7122, with the epoch-290 canonical s2 at 0.7052 as a second witness. The
synthetic-shift picture differs from the natural-shift one. The single fine-tuned IN-C row
(seed 0) scores 0.7548, above the completed distilled cluster by $\sim$4.5pp, and the single
scratch row scores 0.7220, also above that cluster. On this probe the natural-shift ordering
does not hold. We report the asymmetry rather than resolve it: it rests on one fine-tuned and one
scratch seed, IN-C's corruption families interact with the training augmentations differently
across conditions, and no primary claim cites IN-C alone.

\paragraph{Q/K projections under corruption.}
The stability ordering survives one level deeper than attention maps. Clean-corrupt cosines of
the query and key projections are higher for distilled students, at Q 0.795 and K 0.760, than
for fine-tuned students, at Q 0.733 and K 0.729, or scratch students, at Q 0.734 and K 0.736
(all seed means). We report this as
directional evidence only: it locates the stability difference at or before the Q/K linear
maps, and we do not pursue per-layer attribution.

\paragraph{Where the residual gap lives.}
The completed-schedule gap decomposes per suite. Fine-tuning leads on ImageNet-A at every seed,
by +2.5/+3.8/+3.0pp; this is the suite with the widest paired-delta CIs, half-width
$\sim$2.6pp, about 10\% wider than V2 and roughly double Sketch's. Completed distillation
exceeds fine-tuning on ImageNet-R at all three seeds and on Sketch in two of three.
ImageNet-V2 shows small fine-tune leads at every seed, +0.1 to +1.2pp. The residual
completed-schedule gap is therefore not a uniform deficit: fine-tuning's lead concentrates in
IN-A (the widest-interval suite in the paired-delta coordinate) and V2, while distillation
leads on R throughout and Sketch is split. Per-suite values for every model, with CIs, are in
Appendix~\ref{app:suites}.

\begin{table}[b]
\centering
\caption{Secondary synthetic-shift probe: ImageNet-C mean effective robustness on the frozen corruption set, role-labeled. Registered canonical rows plus the registered wave-2 maturity extension.}
\label{tab:incc}
\begin{tabular}{llr}
\toprule
Model & Role & IN-C eff.\ rob. \\
\midrule
distill\_s0 (e145 endpoint) & canonical & 0.6453 \\
finetune\_s0 & canonical & 0.7548 \\
scratch\_s0 & canonical & 0.7220 \\
distill+ADL $\lambda$1.0\_s0 & canonical & 0.7101 \\
distill\_s0 continuation (e300) & maturity contrast & 0.7067 \\
distill\_s1 continuation (e300) & maturity contrast & 0.7098 \\
distill\_s2 continuation (e300) & maturity contrast & 0.7122 \\
distill\_s2 (e290 endpoint) & second witness & 0.7052 \\
distill+ADL $\lambda$3.0\_s0 & dose probe & 0.6948 \\
\bottomrule
\end{tabular}
\end{table}

\section{Discussion}\label{sec:discussion}

Three results triangulate one conclusion. The copy is verified near-perfect and
training-invariant, so the deficit is not a degraded transfer. The gap closes within-seed under
continued training, while the transferred structure never moves, so what was missing was something
training builds slowly, and it is not the routing. And forcing the routing through half a
condition-width of structural change, with accuracy, sharpness, and fidelity held, moves
robustness by nothing detectable, so the visible structure shows no detectable causal effect
on robustness at matched skill in this regime. The account left standing is Li et al.'s own untested conjecture,
now with instrumentation behind it: attention transfer hands the student the teacher's routing
but not the teacher's features, and the robustness the student eventually gains, it grows
itself, slowly, under the transplanted gaze. We state the boundary as plainly as the finding:
this is elimination plus intervention, not a direct exhibition of the features, and its scope
is the regime we measured.

Our results join the ``attention is not explanation'' debate \citep{jain2019attention,
wiegreffe2019attention} from a new axis. That literature asked whether attention explains
predictions. We asked whether attention structure carries robustness, and we ran the transplant
experiment the correlational thread could not: hold the ``good-looking'' attention fixed by
construction, then watch whether the property it supposedly certifies arrives. It does not, and
forcing the structure's most-cited pathology axis to move does not change that. The descriptive
use of attention maps survives untouched: the overlays are accurate about where a model looks.
The evidential use loses its cleanest support. In our regime, the student wearing the teacher's
exact, diverse, corruption-stable gaze is less robust at rule-governed endpoints than the
fine-tuned student, whose attention drifted far from the teacher's toward higher redundancy
and lower corruption-stability. This is the reverse of the ordering attention quality would
predict. Nor does attention quality inversely predict robustness: the scratch student,
whose attention drifted farthest of all, is also the least robust. Across conditions the two
are simply not coupled. The pull of
the evidential reading is strong. When we look at our own overlay figure
(Figure~\ref{fig:overlays}), the teacher-identical maps instinctively read as the better
model, and the data say that reading is backwards here. Attention overlays show where a model looks, not what it knows.

For practitioners of attention distillation, the deficit is, at our
scale, substantially a budget artifact. Distillation's
robustness matures later than its accuracy, so a stopping rule tuned to accuracy flatness
samples distilled models at their least robust moment, and so does a fixed budget matched to
fine-tuning's clock. Three practical corollaries follow, all cheap:
\begin{itemize}
\item Budget distillation schedules generously. Our distilled runs needed their full
  300-epoch horizon to approach parity, 1.0--2.1$\times$ the stops of 145/220/290 epochs where
  accuracy-based stopping cut them.
\item Never compare methods' robustness at a shared fixed budget without checking both
  maturity clocks, because the verdict can invert with training time.
\item Do not select checkpoints by attention aesthetics, which are uninformative about
  robustness here.
\end{itemize}

Our magnitudes are scale-bound, but the mechanism makes a testable claim at Li et al.'s scale: if
ViT-L attention-transfer students at 200 epochs sit early on their own robustness-maturity
curve, then extending distillation budgets there should compress the gap toward fine-tuning,
with attention structure static throughout. If the gap at that scale instead persists at true
completion, the maturity account is bounded to small models and the feature story needs a
scale-dependent component. Either outcome is informative, and the experiment is nothing more
than a budget extension.

One observation invites theory: supervised classification, left free,
made attention \emph{more} redundant in every condition that could drift (fine-tuning
0.4665 to 0.51; scratch to 0.58), while reconstruction pretraining built the diverse,
corruption-stable structure everyone admires. Whatever classification wants from attention, it
is not what reconstruction builds; and whatever robustness needs, our intervention shows it is
not obtained by pushing the redundancy score back down. Both halves are consistent with the
features carrying the load, with attention structure as an expressive but non-binding
byproduct: readable, transplantable, and causally light.

\section{Limitations and design retrospective}\label{sec:limitations}

\paragraph{Scale.}
Everything here is ViT-S on ImageNet-100 with an MAE-S teacher; magnitudes (gaps, redundancy
levels, dose calibrations) are regime-bound and claimed for no other setting. The mechanism-level
claims travel on arguments, not evidence: the KL term binds routing as an optimization constraint rather than
an emergent phenomenon, and MAE's attention character is documented at larger scales, but
portability remains argued, not shown. Our teacher and student are architecture-matched (MAE
ViT-S to ViT-S); concurrent work shows attention transfer itself can fail under teacher--student
architectural mismatch \citep{qin2026attention}, and our claims live in the matched regime,
which is precisely where the robustness question is well-posed. The compensation for scale is resolution. The full-trajectory instrumentation every analysis
rests on costs several hundred dollars at this scale, and roughly two orders of magnitude more
at Li et al.'s. We bought density with size, and the exported prediction
(Section~\ref{sec:discussion}) is the test we could not afford to run.

\paragraph{The stopping rule.}
Our early-stopping rule, standard patience on validation accuracy, turned out to be a
peak-detector deployed on curves that have no peak. The overfit turn never arrives, consistent
with heavy augmentation suppressing it, and gains trickle at the measurement noise floor. The
rule's firing time therefore becomes a lottery: same-condition runs, differing only in seed,
stopped 145 epochs apart. We quantify it (Appendix~\ref{app:lottery}), report every
endpoint with its stopping context, and ground convergence claims in schedule completion and
measured interruption costs rather than the rule's verdicts. The design lesson is plain. On
no-peak curves, complete the horizon and use stopping rules as monitors. Pilot novel
conditions at $n \ge 2$ seeds, since stop-variance is invisible at $n = 1$.

\paragraph{Adjudication timeline.}
The protocol accumulated amendments mid-study; Appendix~\ref{app:prereg} lists the full set
with their triggers. Two bear directly on how results are adjudicated. The first is the cap-hit flatness test,
adopted after fine-tune seed 0 hit both of its horizons' caps, the initial 100-epoch schedule
and its raised 200-epoch replacement; it is post-hoc for
those two runs and disclosed as such, and prospective for every subsequent cap-hit it governs. The
second is the stop-invariant readout for the causal test, registered before any dosed run
existed. We regard such amendment as the correct response to an instrument failing in use,
provided it is prospective for every case it subsequently governs and labeled post-hoc where
it was post-hoc. We document the timeline so reviewers can regard it otherwise if they choose.

\paragraph{Inference boundaries.}
The features account rests on elimination and intervention, and is consistent-with rather than
proven. The three results of Section~\ref{sec:results} constrain where the deficit can
live, but we do not exhibit the deficient features themselves. Three seeds per condition bound our variance estimates; one dataset, one teacher,
one architecture family bound everything else. Bootstrap intervals quantify image-sampling
noise on fixed benchmarks, not run-to-run variation, which the seed bands carry. The
effective-robustness definition is a boundary too: our ratio form and the fitted-baseline
residual of \citet{taori2020measuring} disagree wherever compared models differ in ID accuracy. The
verdicts rest on matched and near-equal-ID comparisons, but the maturity gradient's sign is
definition-dependent (Appendix~\ref{app:residual}).

\FloatBarrier
\bibliography{references}
\bibliographystyle{tmlr}

\appendix
\section{Protocol commitments and amendments}\label{app:prereg}

\textbf{Fixed before any training run.} The following were set before any compute was spent:
\begin{itemize}
\item the three training conditions and three seeds per condition;
\item the single early-stopping rule of Section~\ref{sec:setup} and the per-condition cosine
  horizons;
\item the effective-robustness gap threshold of 0.01, baked into the analysis code;
\item the three hypotheses stated in Section~\ref{sec:setup};
\item both comparison coordinates, endpoint and matched-accuracy, present in the analysis code
  before any result existed;
\item the $\lambda_{\mathrm{kl}} = 36$ setting, from the derivation of
  Appendix~\ref{app:lambda} alone;
\item the teacher quality gate, its mechanism fixed in advance and its floor pinned at launch
  from the measured probe.
\end{itemize}

\textbf{Amended during the study.} Each amendment below responded to an instrument behaving
unexpectedly, and each governed only data that did not yet exist when it was adopted, with the
one exception noted.

\begin{itemize}
\item \emph{Fine-tune horizon, 100 to 200 epochs.} The initial 100-epoch schedule bound
  while validation was still climbing; the pre-declared pilot-calibration clause was
  exercised once, before the full matrix launched, and never again.
\item \emph{Cap-hit flatness reading} (net validation gain over the final 25 epochs against
  a 0.5pp bar). Adopted after the two fine-tune seed-0 cap-hits that prompted it, so it is
  post-hoc for those two runs and prospective for every later cap-hit
  (Appendix~\ref{app:convergence}).
\item \emph{Convergence-reporting policy} (stopping-rule verdicts are operational only;
  convergence claims rest on completed anneals, shown flatness, seed replication, and
  measured interruption cost). Adopted immediately after the study's first flatness-test
  failure, before any subsequent one.
\item \emph{Stop-invariant readout for the causal test} (matched-accuracy primary, with the
  iso-level variant). Registered after the baseline stop-timing variance was measured and
  before any dosed run existed; the iso-level specification was added before any dosed
  evaluation.
\item \emph{Full-schedule rerun program.} Registered before the matrix completed, with an
  escalation criterion (extend beyond seed 0 iff the seed-0 delta reached 1.0pp ID or 0.01
  effective robustness) fixed before the first rerun existed; extended to all three seeds
  before any rerun data existed, superseding the criterion, which the seed-0 result then
  also satisfied.
\item \emph{Wave-2 ImageNet-C extension.} The additional role-labeled rows of
  Table~\ref{tab:incc} were registered before those evaluations existed, and extended to
  all three rerun endpoints when the rerun program grew.
\item \emph{Verdict-interpretation table.} The outcome-to-conclusion mapping for both
  verdicts was frozen before any verdict-bearing data existed; Sections~\ref{sec:gap}
  and~\ref{sec:causal} report their results against it.
\end{itemize}

\textbf{One in-place correction.} The registration record was edited in place exactly once:
the note accepting the $\lambda$1.0 s1 cap-hit described that run's trajectory tail
incorrectly, and the description was corrected and labeled at the site
(Appendix~\ref{app:convergence} carries the corrected numbers). Every other change over the
study's life was an addition, and no rule, threshold, or reported result was ever edited.
The registry and the project's working records are retained; this appendix is the complete
list of protocol-relevant commitments and changes.

\textbf{The second hypothesis.} The prediction that the layer-wise overfocusing
signature would predict the ID-to-OOD gap is reported nowhere as a fitted correlation, and
the reason is recorded here. The measurements left the registered form unanswerable rather
than answered: attention structure proved near-invariant within condition
(Section~\ref{sec:faithful}) while the gap proved highly variable with stopping time
(Section~\ref{sec:gap}), so an observational fit would relate a predictor taking three
effective values, one per condition and collinear with it, to an outcome dominated by
training maturity. The dose ladder of Section~\ref{sec:causal} tests the same linkage with
the predictor manipulated within condition instead, and Section~\ref{sec:discussion} reports
the cross-condition observation in prose: the two are not coupled.

\section{The $\lambda_{\mathrm{kl}} = 36$ derivation and gradient-identity check}\label{app:lambda}

Li et al.'s released implementation computes the attention-transfer loss as a teacher-as-target
cross-entropy over attention maps: gradient-identical to a KL divergence, since the
teacher-entropy term is constant in the student, though offset in value. It is averaged over
batch, heads, and query rows, then summed over the distilled layers, scaled by $\lambda = 3$.
Our implementation averages over layers as well, a mean of means, which changes the effective
per-layer weight. Matching Li et al.'s per-layer gradient scale at 12 distilled layers therefore
requires $\lambda_{\mathrm{kl}} = 3 \times 12 = 36$. This is arithmetic, not tuning: the two
formulations are identical up to the constant. We verified the identity directly. Autograd
gradients of the two loss formulations, $36 \times$ mean-over-12-layers KL against $3 \times$
the sum over 12 layers of Li et al.'s verbatim loss, were evaluated on identical randomly-drawn
attention logits and agree to a maximum absolute difference of $5.6 \times 10^{-9}$, which is
float32 noise. This is a loss-level gradient identity on synthetic inputs, which is the
quantity the derivation asserts.

Two deliberate loss-side departures from Li et al.'s distillation recipe are disclosed here.
Training-protocol differences such as budgets, stopping rule, and teacher provenance are
Section~\ref{sec:setup}'s subject. First, we distill all 12 layers rather than their default
first-18-of-24, for treatment purity: the analyses then measure the same quantity the loss
optimizes, at every layer. Second, our batch size and learning-rate schedule follow the
standard DeiT-style recipe at our scale rather than their ViT-L configuration. Our design rule
was to be faithful to the method, not the numbers. The $\lambda_{\mathrm{kl}} = 36$ value was pinned before the matrix
ran, from this derivation alone; the pilot's transfer-bound check (the KL term collapsing
$1.67 \to 0.02$ and flattening) confirmed it binds as intended.

\section{Convergence adjudication records}\label{app:convergence}

Every run ends in one of three recorded states: early-stop certificate (the rule fired),
cap-hit with a passing terminal-flatness reading, or cap-hit with a failing one, which is
never auto-accepted and always adjudicated on the record. The flatness reading is the net
validation top-1 gain over the final 25 epochs against a 0.5pp bar; it is descriptive evidence,
never a convergence certificate (per-eval noise is $\pm$0.44pp, so single-interval deltas are
sub-noise by construction). Three cap-hits failed the bar and were accepted with disclosure. All three overages are
small, 0.02--0.08pp over the 0.5pp bar, on fully-annealed tails whose per-eval gaps sit at
or below the $\pm$0.44pp noise. None shows the consecutive above-noise climb of the one
trend-shaped case observed, at +0.205pp/eval. The discriminator separates the classes by
roughly 2$\times$ in rate and 4--8$\times$ in overage.

\begin{itemize}
\item \textbf{finetune s2}: net +0.52pp vs.\ the 0.50 bar (fail by 0.02pp, one image);
  final five gaps +0.12/+0.12/+0.24/+0.06/$-$0.02, each at or below per-eval noise. Accepted under the
  completed-schedule policy with family-band evidence: the three fine-tuned seeds ended via
  three different formal outcomes (a rule-issued stop certificate, a certificate completed
  only at the horizon's final evaluation, and this failed flatness reading) yet landed in
  a 0.3pp ID band (0.8896--0.8926), effective robustness 0.4913--0.4964. The formalism's
  verdicts are demonstrably uncorrelated with the answers.
\item \textbf{$\lambda$0.3 s0}: net +0.58pp (fail by 0.08); final gaps
  +0.16/+0.02/+0.38/$-$0.30/+0.32, the largest at epoch 289. Accepted per the same precedent.
\item \textbf{$\lambda$1.0 s1}: net +0.54pp (fail by 0.04); final gaps
  +0.24/+0.14/+0.24/$-$0.12/+0.04, three positive: the least clean of the three accepts.
  Accepted on the quantitative discriminator (rate 0.108pp/eval, roughly half the one
  trend-shaped case ever observed; overage an eighth of that case's).
\end{itemize}

The discriminator separating noise-shaped from trend-shaped overage was validated by the one
trend-shaped case: the original 100-epoch fine-tune horizon failed with +0.205pp/eval while
still climbing, and its extension recovered +2.0pp (0.8718 $\to$ 0.8916 best validation
top-1), retro-verifying the classification. The
false-fail mode of the flatness test is single-jitter aliasing, quantified above. Full
validation trajectories for every run are published with the artifacts.

\section{Continuation program (full-schedule reruns)}\label{app:continuation}

\textbf{Registration.} The seed-0 rerun was registered before the matrix completed, with a
pinned escalation criterion fixed before the run existed: extend to seeds 1--2 iff the
seed-0 delta reached 1.0pp ID or 0.01 effective robustness. The criterion fired, at +6.0pp ID and
+0.053 effective robustness. Before any continuation data existed the program had been amended
to all three seeds regardless (Appendix~\ref{app:prereg}). That made seed 2 an in-design
negative control with a predicted near-zero delta, since it had been cut only 10 epochs from
its horizon, at 1.3\% of peak learning rate.

\textbf{Instrument validation.} Same-seed replay is exact in this stack: across 131 shared
evaluations spanning the three seeds, the reruns reproduce their canonical prefixes with a
maximum absolute validation difference of 0.0. Each rerun is therefore the canonical run
continued, not a new draw. The control seed moved +0.28pp ID / +0.71pp effective robustness,
below the 1pp instrument-suspicion bar, and its nonzero remainder is the genuinely remaining
10 epochs of schedule.

\textbf{Readings.} Interruption cost is monotone in how hot the rule cut: 59\% of peak
learning rate at the cut recovered +6.02pp ID / +5.31pp effective robustness; 20\% recovered
+1.76 / +3.11pp; 1.3\% of peak (the schedule's floor plus 0.3\% of remaining decay)
recovered +0.28 / +0.71pp. Effective robustness outgained ID accuracy in the two later
seeds' windows; seed 0's full remainder moved both together (+5.31 vs.\ +6.02pp). Structure
at the completed schedules' promoted checkpoints stayed at the teacher's value in all three
seeds
(redundancy 0.4673/0.4676/0.4673 against the teacher's 0.4665). Continuations are labeled
artifacts with first-class evidential use and second-class authority: registered claims are
computed from rule-endpoints alone, and no continuation number enters a primary table
unlabeled.

\section{Stop-lottery statistics}\label{app:lottery}

\begin{figure}[h]
\centering
\includegraphics[width=0.9\textwidth]{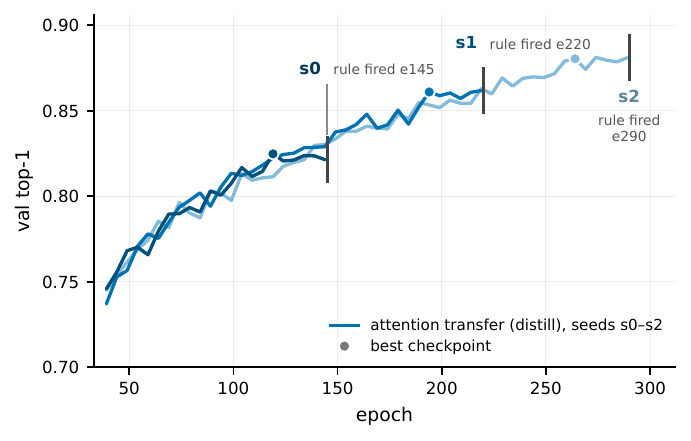}
\caption{Validation trajectories of the three distilled seeds under the single
pre-registered stopping rule, with rule-fire epochs and best checkpoints
marked; line shade distinguishes seeds. Near-identical no-peak curves; stops 145 epochs apart.}
\label{fig:lottery}
\end{figure}

The distill family exposes the stopping rule's variance. Under one rule and one configuration,
three seeds stopped at epochs 145, 220, and 290 (Figure~\ref{fig:lottery}), giving an ID spread
of 5.6pp and an effective-robustness spread of 4.1pp at the endpoints. The
attention-structure metrics meanwhile replicated tightly across the same runs: redundancy
$\pm$0.0008, top-$k$ $\pm$0.0011, entropy $\pm$0.006, and stability $\pm$0.002, each one to
three orders of magnitude below its between-condition separation. The curves themselves are
near-identical: seed 1 at epoch 144 sat at 0.8290, within noise of seed 0's final 0.8248; the
difference is where 25 flat epochs happened to align with the $\pm$0.44pp per-eval noise. The
rule is a peak-detector and these curves have no peak, so its firing time is a lottery over
noise alignments, and endpoint metrics inherit that lottery. Section~\ref{sec:gap}'s
$r = 0.978$ is the same fact viewed across the family. Fine-tuning and scratch, whose plateaus
fit within their horizons, show the ordinary version: seed bands of 0.3--0.5pp on both
metrics. Fine-tuning spans ID 0.8896--0.8926 and effective robustness 0.4913--0.4964; scratch
spans 0.8462--0.8498 and 0.4168--0.4199.

\section{Full per-suite tables and matched-pair selections}\label{app:suites}

Table~\ref{tab:persuite} reports per-suite effective robustness for all 24 endpoint models with
bootstrap CIs on the 4-suite mean. Table~\ref{tab:selections} lists every matched-pair
checkpoint selection used in Sections~\ref{sec:matched} and \ref{sec:causal}: run, target
level, selected epoch, achieved validation accuracy, and residual. Residuals are at most 0.0034
for every dose and iso-level selection; the one exception is the seed-1 own-level ladder
rung, whose fine-tuned side sits 0.0054 below target and is machine-flagged loose against
the 0.005 tolerance. Within-pair ID differences are at most 0.22pp in the dose analyses,
0.36pp in the iso-level analyses, and 0.54pp at that one ladder rung.

\begin{table}[t]
\centering\small
\caption{Per-suite effective robustness for every evaluated endpoint model. Mean column carries the paired bootstrap 95\% CI.}
\label{tab:persuite}
\begin{tabular}{lrrrrl}
\toprule
Model & IN-A & IN-R & IN-Sk & IN-V2 & Mean [95\% CI] \\
\midrule
scratch\_vit\_s\_s0 & 0.1088 & 0.3871 & 0.3199 & 0.8637 & 0.4199 [0.4070, 0.4332] \\
scratch\_vit\_s\_s1 & 0.1196 & 0.3919 & 0.3107 & 0.8451 & 0.4168 [0.4038, 0.4301] \\
scratch\_vit\_s\_s2 & 0.1303 & 0.3922 & 0.3122 & 0.8402 & 0.4188 [0.4053, 0.4324] \\
finetune\_vit\_s\_s0 & 0.1453 & 0.4953 & 0.4405 & 0.8961 & 0.4943 [0.4814, 0.5073] \\
finetune\_vit\_s\_s1 & 0.1564 & 0.4805 & 0.4302 & 0.8982 & 0.4913 [0.4785, 0.5044] \\
finetune\_vit\_s\_s2 & 0.1433 & 0.4965 & 0.4440 & 0.9019 & 0.4964 [0.4836, 0.5092] \\
distill\_vit\_s\_s0 & 0.0829 & 0.4195 & 0.3964 & 0.8487 & 0.4369 [0.4233, 0.4510] \\
distill\_vit\_s\_s1 & 0.0913 & 0.4635 & 0.4101 & 0.8711 & 0.4590 [0.4461, 0.4720] \\
distill\_vit\_s\_s2 & 0.1034 & 0.4872 & 0.4492 & 0.8735 & 0.4783 [0.4658, 0.4910] \\
distill\_vit\_s\_s0\_fullsched & 0.1204 & 0.4998 & 0.4447 & 0.8949 & 0.4900 [0.4773, 0.5023] \\
distill\_vit\_s\_s1\_fullsched & 0.1184 & 0.5023 & 0.4473 & 0.8923 & 0.4901 [0.4774, 0.5030] \\
distill\_vit\_s\_s2\_fullsched & 0.1137 & 0.4972 & 0.4409 & 0.8899 & 0.4854 [0.4727, 0.4980] \\
distill\_h3\_vit\_s\_lam0.1\_s0 & 0.0876 & 0.4349 & 0.3886 & 0.8464 & 0.4394 [0.4261, 0.4532] \\
distill\_h3\_vit\_s\_lam0.1\_s1 & 0.1227 & 0.4983 & 0.4461 & 0.8932 & 0.4901 [0.4772, 0.5032] \\
distill\_h3\_vit\_s\_lam0.1\_s2 & 0.0887 & 0.4471 & 0.3897 & 0.8725 & 0.4495 [0.4367, 0.4628] \\
distill\_h3\_vit\_s\_lam0.3\_s0 & 0.1127 & 0.4942 & 0.4474 & 0.8885 & 0.4857 [0.4731, 0.4984] \\
distill\_h3\_vit\_s\_lam0.3\_s1 & 0.1182 & 0.4916 & 0.4431 & 0.8829 & 0.4840 [0.4713, 0.4967] \\
distill\_h3\_vit\_s\_lam0.3\_s2 & 0.1200 & 0.5022 & 0.4546 & 0.8954 & 0.4931 [0.4805, 0.5061] \\
distill\_h3\_vit\_s\_lam1.0\_s0 & 0.1260 & 0.4836 & 0.4360 & 0.9024 & 0.4870 [0.4743, 0.5000] \\
distill\_h3\_vit\_s\_lam1.0\_s1 & 0.1229 & 0.4871 & 0.4526 & 0.8857 & 0.4871 [0.4743, 0.4999] \\
distill\_h3\_vit\_s\_lam1.0\_s2 & 0.0969 & 0.4529 & 0.4064 & 0.8897 & 0.4615 [0.4486, 0.4746] \\
distill\_h3\_vit\_s\_lam3.0\_s0 & 0.1048 & 0.4923 & 0.4465 & 0.8698 & 0.4783 [0.4656, 0.4911] \\
distill\_h3\_vit\_s\_lam3.0\_s1 & 0.1056 & 0.4853 & 0.4486 & 0.8806 & 0.4800 [0.4674, 0.4929] \\
distill\_h3\_vit\_s\_lam3.0\_s2 & 0.1211 & 0.5016 & 0.4527 & 0.8863 & 0.4904 [0.4777, 0.5034] \\
\bottomrule
\end{tabular}
\end{table}

\begin{table}[h]
\centering\scriptsize
\caption{Every matched-pair checkpoint selection: analysis, model, achieved validation top-1, selected epoch, and residual to the target level. All residuals $\le 0.0034$ except the s1 own-level ladder rung's fine-tuned side ($-$0.0054, machine-flagged loose against the 0.005 tolerance).}
\label{tab:selections}
\begin{tabular}{llrrr}
\toprule
Analysis & Model & Val top-1 & Epoch & Residual \\
\midrule
dose per-pair $\lambda$0.1 & sweep s0 & 0.8234 & 129 & -0.0014 \\
dose per-pair $\lambda$0.1 & baseline s0 & 0.8248 & 119 & +0.0000 \\
dose per-pair $\lambda$0.1 & sweep s1 & 0.8622 & 219 & +0.0004 \\
dose per-pair $\lambda$0.1 & baseline s1 & 0.8618 & 219 & +0.0000 \\
dose per-pair $\lambda$0.1 & sweep s2 & 0.8596 & 189 & +0.0000 \\
dose per-pair $\lambda$0.1 & baseline s2 & 0.8598 & 224 & +0.0002 \\
dose per-pair $\lambda$0.3 & sweep s0 & 0.8242 & 139 & -0.0006 \\
dose per-pair $\lambda$0.3 & baseline s0 & 0.8248 & 119 & +0.0000 \\
dose per-pair $\lambda$0.3 & sweep s1 & 0.8612 & 219 & -0.0006 \\
dose per-pair $\lambda$0.3 & baseline s1 & 0.8618 & 219 & +0.0000 \\
dose per-pair $\lambda$0.3 & sweep s2 & 0.8818 & 274 & +0.0006 \\
dose per-pair $\lambda$0.3 & baseline s2 & 0.8812 & 274 & +0.0000 \\
dose per-pair $\lambda$1.0 & sweep s0 & 0.8228 & 129 & -0.0020 \\
dose per-pair $\lambda$1.0 & baseline s0 & 0.8248 & 119 & +0.0000 \\
dose per-pair $\lambda$1.0 & sweep s1 & 0.8622 & 209 & +0.0004 \\
dose per-pair $\lambda$1.0 & baseline s1 & 0.8618 & 219 & +0.0000 \\
dose per-pair $\lambda$1.0 & sweep s2 & 0.8582 & 209 & +0.0000 \\
dose per-pair $\lambda$1.0 & baseline s2 & 0.8598 & 224 & +0.0016 \\
dose per-pair $\lambda$3.0 & sweep s0 & 0.8246 & 144 & -0.0002 \\
dose per-pair $\lambda$3.0 & baseline s0 & 0.8248 & 119 & +0.0000 \\
dose per-pair $\lambda$3.0 & sweep s1 & 0.8596 & 224 & -0.0022 \\
dose per-pair $\lambda$3.0 & baseline s1 & 0.8618 & 219 & +0.0000 \\
dose per-pair $\lambda$3.0 & sweep s2 & 0.8808 & 274 & -0.0004 \\
dose per-pair $\lambda$3.0 & baseline s2 & 0.8812 & 274 & +0.0000 \\
iso-level ($L^{*}=0.8248$) & sweep $\lambda$0.1 s0 & 0.8234 & 129 & -0.0014 \\
iso-level ($L^{*}=0.8248$) & sweep $\lambda$0.1 s1 & 0.8230 & 124 & -0.0018 \\
iso-level ($L^{*}=0.8248$) & sweep $\lambda$0.1 s2 & 0.8250 & 139 & +0.0002 \\
iso-level ($L^{*}=0.8248$) & sweep $\lambda$0.3 s0 & 0.8242 & 139 & -0.0006 \\
iso-level ($L^{*}=0.8248$) & sweep $\lambda$0.3 s1 & 0.8272 & 149 & +0.0024 \\
iso-level ($L^{*}=0.8248$) & sweep $\lambda$0.3 s2 & 0.8242 & 134 & -0.0006 \\
iso-level ($L^{*}=0.8248$) & sweep $\lambda$1.0 s0 & 0.8228 & 129 & -0.0020 \\
iso-level ($L^{*}=0.8248$) & sweep $\lambda$1.0 s1 & 0.8248 & 134 & +0.0000 \\
iso-level ($L^{*}=0.8248$) & sweep $\lambda$1.0 s2 & 0.8248 & 139 & +0.0000 \\
iso-level ($L^{*}=0.8248$) & sweep $\lambda$3.0 s0 & 0.8246 & 144 & -0.0002 \\
iso-level ($L^{*}=0.8248$) & sweep $\lambda$3.0 s1 & 0.8254 & 129 & +0.0006 \\
iso-level ($L^{*}=0.8248$) & sweep $\lambda$3.0 s2 & 0.8234 & 124 & -0.0014 \\
iso-level ($L^{*}=0.8248$) & baseline s0 & 0.8248 & 119 & +0.0000 \\
iso-level ($L^{*}=0.8248$) & baseline s1 & 0.8244 & 124 & -0.0004 \\
iso-level ($L^{*}=0.8248$) & baseline s2 & 0.8214 & 134 & -0.0034 \\
ladder top s0 & distill side & 0.8850 & 294 & +0.0000 \\
ladder top s0 & ft side & 0.8844 & 164 & -0.0006 \\
ladder top s1 & distill side & 0.8794 & 299 & +0.0000 \\
ladder top s1 & ft side & 0.8806 & 149 & +0.0012 \\
ladder top s2 & distill side & 0.8832 & 299 & +0.0000 \\
ladder top s2 & ft side & 0.8802 & 149 & -0.0030 \\
ladder own-level s0 & distill side & 0.8248 & 119 & +0.0000 \\
ladder own-level s0 & ft side & 0.8240 & 69 & -0.0008 \\
ladder own-level s1 & distill side & 0.8618 & 219 & +0.0000 \\
ladder own-level s1 & ft side & 0.8564 & 104 & -0.0054 \\
ladder own-level s2 & distill side & 0.8812 & 274 & +0.0000 \\
ladder own-level s2 & ft side & 0.8802 & 149 & -0.0010 \\
\bottomrule
\end{tabular}
\end{table}

\FloatBarrier
\section{Residual-form sensitivity check}\label{app:residual}

We report effective robustness as the ratio OOD/ID on covered classes
(Section~\ref{sec:setup}); \citet{taori2020measuring} and \citet{miller2021accuracy} define
it as the residual of OOD accuracy above a baseline fit to ID accuracy in a scaled space
(log-linear in Taori et al., probit in Miller et al.; our check follows Miller). The two
definitions order models identically wherever compared models sit at matched or near-equal ID,
which is where both of this paper's verdicts live. They can disagree where ID differs, which
covers two reported quantities: the endpoint-coordinate gap triple and the maturity
correlation. Table~\ref{tab:residual} recomputes both under the residual form for three
baseline-population choices. This check is post-hoc: it was computed at final review and is not part of the
pre-registered analysis set.

\begin{table}[h]
\centering\small
\caption{Residual-form sensitivity check (post-hoc, computed at final review; not pre-registered). Per suite, probit(OOD) is fit linearly on probit(covered-ID) over the stated baseline population; a model's score is its mean residual over the four suites (probit units). Ratio-form references: $r = +0.978$; endpoint gaps +0.0575 / +0.0323 / +0.0181.}
\label{tab:residual}
\begin{tabular}{lrrrr}
\toprule
Baseline population & $r$(stop, residual) & \multicolumn{3}{c}{ft $-$ distill endpoint residual gap} \\
\cmidrule(lr){3-5}
 & & s0 & s1 & s2 \\
\midrule
all 24 endpoint models & -0.562 & -0.0181 & -0.0122 & +0.0525 \\
fine-tune + scratch & -0.854 & -0.0841 & -0.0464 & +0.0421 \\
ft + scratch + dose-zero distill & -0.689 & -0.0335 & -0.0255 & +0.0515 \\
\bottomrule
\end{tabular}
\end{table}

Under every non-degenerate baseline population the residual-form maturity correlation is
negative, $-$0.56 to $-$0.85, where the ratio form reads +0.978. Later distilled checkpoints
gain OOD accuracy, but more slowly than the cross-model probit trend predicts from their ID
gains. The endpoint residual gaps also change sign for seeds 0 and 1. Neither movement touches
the verdicts, which rest on matched- and near-equal-ID comparisons, but it means the maturity
gradient and the endpoint triple are ratio-form statements, and readers in the residual
tradition should read them as such (the main text labels them so). An in-study baseline of 24
models spanning 0.825--0.893 ID under-determines a Taori-style fit, so this is a sensitivity
check, not a replacement measurement; the direction is nonetheless stable across every
non-degenerate fit we tried.

\section{Training configuration and compute}\label{app:config}

All runs share one training stack, resolved from version-controlled configuration files; the
fully resolved configuration ships with every run's artifacts. The classifier recipe is the
standard DeiT-style setup at this scale: AdamW with weight decay 0.05, batch size 256,
cosine learning-rate decay to a floor of $10^{-5}$, gradient clipping at 1.0, label smoothing
0.1, stochastic depth 0.1, mixed-precision training, and the DeiT augmentation stack
(RandAugment m9, mixup 0.8, cutmix 1.0, color jitter 0.3, random erasing 0.25). Validation
and checkpointing run every 5 epochs (Section~\ref{sec:setup}). Checkpoints are stored in
full precision.

\begin{table}[h]
\centering
\caption{Per-condition training settings; all other hyperparameters are shared as above.}
\label{tab:config}
\begin{tabular}{lrrr}
\toprule
 & Peak LR & Warmup (epochs) & Horizon (epochs) \\
\midrule
Scratch, distill, and dose arms & $10^{-3}$ & 20 & 300 \\
Fine-tune & $5 \times 10^{-4}$ & 5 & 200 \\
MAE teacher (pretraining) & $1.5 \times 10^{-4}$ & 40 & 400 (probe-stopped) \\
\bottomrule
\end{tabular}
\end{table}

The MAE teacher additionally uses mask ratio 0.75 with a 256-dimensional, 4-layer decoder
and normalized-pixel loss, decaying to a $10^{-6}$ floor. All runs used one NVIDIA
A100-SXM4-40GB (PyTorch 2.12.1, CUDA 12.6, timm 1.0.27, Python 3.10); training compute,
summed from per-run logs, was 282.5 GPU-hours over the 24 classifier runs (9 core, 12
dosed, 3 reruns) plus 8.7 for the teacher, about 291 in all.

\end{document}